\documentclass[]{style}

\definecolor{lightblue}{rgb}{0.0, 0.6, 1.0}
\definecolor{darkgreen}{rgb}{0.0, 0.6, 0.0}
\definecolor{lightgreen}{rgb}{0.1, 0.8, 0.1}
\definecolor{lightred}{rgb}{0.7, 0.7, 0.7}
\definecolor{lightgreen}{rgb}{0, 0, 0}
\definecolor{lightlightgray}{rgb}{0.8, 0.8, 0.8}

\usepackage{titlesec}
\titlespacing*{\paragraph}
{0pt}{0em}{0.2em}  

\makeatletter
\renewcommand\paragraph{\@startsection{paragraph}{4}{\z@}%
  {3pt}
  {-0.5em}
  {\normalfont\normalsize\bfseries}} 
\makeatother

\usepackage{amsmath}  
\usepackage{upgreek}
\usepackage{array}    
\usepackage{multirow}
\usepackage{array}
\usepackage{algorithm}
\usepackage{makecell}
\usepackage{graphicx}
\usepackage{algorithm}
\usepackage{algpseudocode}
\usepackage{colortbl}
\usepackage{enumitem}
\usepackage{wrapfig}
\usepackage{mathrsfs}
\usepackage{pifont}
\usepackage{soul}
\usepackage{amssymb}
\usepackage{tcolorbox}
\usepackage{dirtytalk}
\usepackage{xspace}
\usepackage[utf8]{inputenc}

\usepackage{booktabs}

\usepackage{url}
\definecolor{blue}{rgb}{0.21,0.49,0.74}
\definecolor{red}{rgb}{0.8, 0.2, 0.2}
\definecolor{green}{rgb}{0, 0.5, 0}
\definecolor{yellow}{RGB}{218, 160, 109}
\definecolor{gray}{RGB}{155, 155, 155}

\crefname{section}{Sec.}{Secs.}
\Crefname{section}{Section}{Sections}
\Crefname{table}{Table}{Tables}
\crefname{table}{Tab.}{Tabs.}
\crefname{figure}{Fig.}{Figs.}
\Crefname{figure}{Figure}{Figures}
\crefname{appendix}{App.}{Apps.}
\Crefname{appendix}{Appendix}{Appendices}

\newcommand{\GRs}{\texttt{VEFX-Reward-4B}\xspace}
\newcommand{\GRl}{\texttt{VEFX-Reward-32B}\xspace}
\newcommand{\GR}{\texttt{VEFX-Reward}\xspace}
\newcommand{\GB}{\texttt{VEFX-Bench}\xspace}
\newcommand{\GD}{\texttt{VEFX-Dataset}\xspace}
\newcommand{\ldash}{\textcolor{lightred}{--}}
\newcommand{\lX}{\textcolor{lightred}{\xmark}}
\newcommand{\CK}{\textcolor{lightgreen}{\cmark}}

\usepackage[utf8]{inputenc} 
\usepackage[T1]{fontenc}    
\usepackage{graphicx}
\usepackage{amsmath}
\usepackage{amssymb}
\usepackage[numbers,sort&compress]{natbib}
\usepackage{url}            
\usepackage{booktabs}       
\usepackage{amsfonts}       
\usepackage{nicefrac}       
\usepackage{microtype}      
\usepackage{xcolor}         
\usepackage{multirow}
\usepackage{comment}
\usepackage{colortbl}
\usepackage{tocloft}
\usepackage{algorithm}
\usepackage{algorithmicx}
\usepackage{algpseudocode}
\usepackage{wrapfig}
\usepackage{tabularx}
\usepackage{textcomp}
\usepackage{stfloats}
\usepackage{verbatim}
\usepackage{titlesec}
\usepackage{adjustbox}
\usepackage{pifont}
\usepackage{tikz}
\usepackage{color}
\usepackage{subcaption}
\usepackage{soul}

\RequirePackage{xspace}
\makeatletter
\DeclareRobustCommand\onedot{\futurelet\@let@token\@onedot}
\def\@onedot{\ifx\@let@token.\else.\null\fi\xspace}

\makeatother

\definecolor{lightblue}{rgb}{0.66, 0.85, 0.95}
\hypersetup{
    colorlinks=true,
    linkcolor=red,
    citecolor=blue,
}
\definecolor{c2}{HTML}{FBD9BD}
\definecolor{c3}{HTML}{fe793d}
\definecolor{c4}{HTML}{eedeb0}
\definecolor{rouse}{rgb}{0.981,0.961,0.941}
\usepackage[most]{tcolorbox}
\tcbset{colback=blue!5!white, colframe=blue!20!white, boxrule=0pt, arc=2mm, left=1mm, right=1mm, top=1mm, bottom=1mm}

\definecolor{adptorange}{RGB}{248, 205, 172}
\definecolor{cmpblue}{RGB}{189, 215, 238}

\definecolor{our_red}{RGB}{232,157,160}
\definecolor{our_blue}{RGB}{136,206,230}
\definecolor{our_orange}{RGB}{246,200,168}
\definecolor{our_green}{RGB}{178,211,164}

\definecolor{token_blue}{RGB}{84, 120, 140}

\usepackage{pifont}       
\usepackage{bbding}
\usepackage{fontawesome}
\usepackage{xspace}
\newcommand{\xmark}{\ding{55}}
\newcommand{\cmark}{\ding{51}}
\usepackage{float}

\newlength\savewidth\newcommand\shline{\noalign{\global\savewidth\arrayrulewidth \global\arrayrulewidth 1pt}\hline\noalign{\global\arrayrulewidth\savewidth}}
\newcommand{\tablestyle}[2]{\setlength{\tabcolsep}{#1}\renewcommand{\arraystretch}{#2}\centering\footnotesize}

\newcolumntype{x}[1]{>{\centering\arraybackslash}p{#1pt}}
\newcolumntype{y}[1]{>{\raggedright\arraybackslash}p{#1pt}}
\newcolumntype{z}[1]{>{\raggedleft\arraybackslash}p{#1pt}}

\renewcommand{\paragraph}[1]{\vspace{1.25mm}\noindent\textbf{#1}}

\usepackage{colortbl}
\usepackage{wrapfig}

\usepackage{listings}

\definecolor{codeblue}{rgb}{0.21, 0.49, 0.74}
\definecolor{codekw}{rgb}{0.35, 0.35, 0.75}
\lstdefinestyle{Pytorch}{
    language = Python,
    backgroundcolor = \color{white},
    basicstyle = \fontsize{9pt}{8pt}\selectfont\ttfamily\bfseries,
    columns = fullflexible,
    aboveskip=1pt,
    belowskip=1pt,
    breaklines = true,
    captionpos = b,
    commentstyle = \color{codeblue},
    keywordstyle = \color{codekw},
}

\definecolor{green}{HTML}{009000}
\definecolor{red}{HTML}{ea4335}

\definecolor{G1}{HTML}{7E0B0B}
\definecolor{G2}{HTML}{A31212}
\definecolor{G3}{HTML}{C61C1C}
\definecolor{G4}{HTML}{E50914}

\definecolor{B1}{HTML}{8A5A1F}   
\definecolor{B2}{HTML}{9C6824}   
\definecolor{B3}{HTML}{AD7630}   
\definecolor{B4}{HTML}{BE853D}   
\definecolor{B5}{HTML}{CC944A}   
\definecolor{B6}{HTML}{D8A45A}   

\newcommand{\VEFXBenchLuxury}{%
{
\textcolor{G1}{V}%
\textcolor{G2}{E}%
\textcolor{G3}{F}%
\textcolor{G4}{X}%
\textcolor{B1}{-}%
\textcolor{B2}{B}%
\textcolor{B3}{e}%
\textcolor{B4}{n}%
\textcolor{B5}{c}%
\textcolor{B6}{h}}%
}

\title{%
\VEFXBenchLuxury: A Holistic Benchmark for Generic \\
\vspace{2mm}
Video Editing and Visual Effects}

\author[1,2]{Xiangbo Gao}
\author[3]{Sicong Jiang}
\author[3]{Bangya Liu}
\author[1]{Xinghao Chen}
\author[3]{Minglai Yang}
\author[1]{Siyuan Yang}
\author[1]{Mingyang Wu}
\author[1]{Jiongze Yu}
\author[2]{Qi Zheng}
\author[2]{Haozhi Wang}
\author[]{Jiayi Zhang}
\author[2]{Jie Yang}
\author[3]{Zihan Wang}
\author[2]{Qing Yin}
\author[1,2]{Zhengzhong Tu}

\affiliation[1]{Texas A\&M University}
\affiliation[2]{Visko Platform}
\affiliation[3]{Abaka AI}

\abstract{
As AI-assisted video creation becomes increasingly practical, instruction-guided video editing has become essential for refining generated or captured footage to meet professional requirements. Yet the field still lacks both a large-scale human-annotated dataset with complete editing examples and a standardized evaluator for comparing editing systems. Existing resources are limited by small scale, missing edited outputs, or the absence of human quality labels, while current evaluation often relies on expensive manual inspection or generic vision-language model judges that are not specialized for editing quality. We introduce \GD, a human-annotated dataset containing 5,049 video editing examples across 9 major editing categories and 32 subcategories, each labeled along three decoupled dimensions: Instruction Following, Rendering Quality, and Edit Exclusivity. Building on \GD, we propose \GR, a reward model designed specifically for video editing quality assessment. \GR jointly processes the source video, the editing instruction, and the edited video, and predicts per-dimension quality scores via ordinal regression. We further release \GB, a benchmark of 300 curated video-prompt pairs for standardized comparison of editing systems. Experiments show that \GR aligns more strongly with human judgments than generic VLM judges and prior reward models on both standard IQA/VQA metrics and group-wise preference evaluation. Using \GR as an evaluator, we benchmark representative commercial and open-source video editing systems, revealing a gap between visual plausibility, instruction following, and edit locality in current models.
}

\project{\url{https://xiangbogaobarry.github.io/VEFX-Bench/}}
\date{\today}
\contact{Xiangbo Gao (\href{mailto:xiangbogaobarry@gmail.com}{xiangbogaobarry@gmail.com}), Zhengzhong Tu}

\begin{document}
\thispagestyle{firstheader}
\maketitle
\pagestyle{plain}

\begin{figure*}[t]
\centering
\includegraphics[width=\linewidth]{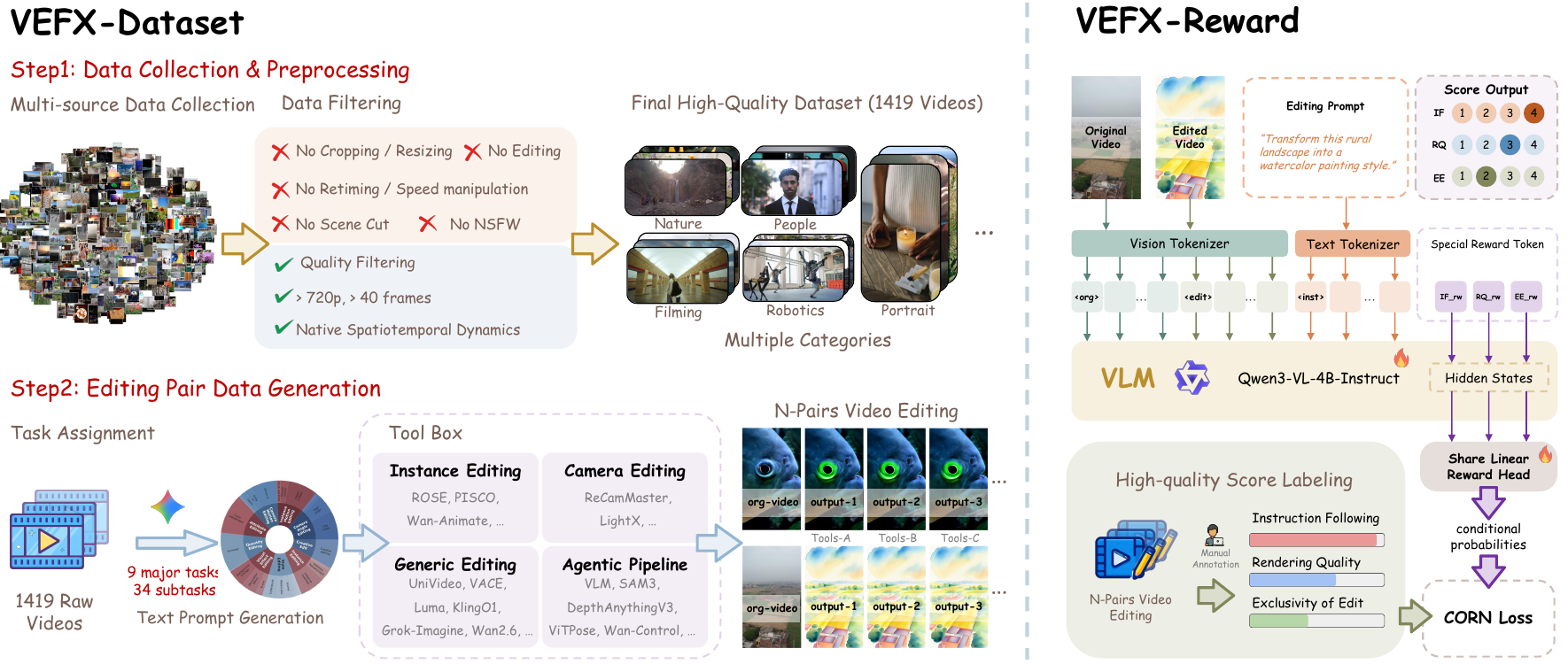}
\caption{\textbf{Overview of our framework.} We construct \GD, a human-annotated dataset with 5,049 video editing examples across 9 categories and 32 subcategories, scored along three decoupled dimensions: Instruction Following (IF), Rendering Quality (RQ), and Edit Exclusivity (EE). We train \GR, a dedicated reward model for video editing quality assessment that takes the original video, editing instruction, and edited video as input and predicts per-dimension quality scores. We further release \GB, a benchmark of 300 curated video-prompt pairs for standardized comparison of editing systems.}
\label{fig:teaser}
\vspace{-0.3cm}
\end{figure*}

\section{Introduction}
\label{sec:intro}

The landscape of AI-assisted video creation is advancing rapidly. Recent video generation systems have shown impressive progress in producing photorealistic clips from natural-language prompts \citep{wan2025wan, chen2025goku, kong2024hunyuanvideo, polyak2024movie, openai_sora_2024, li20244k4dgen, wu2026consid, deepmind_veo3_2025, klingai_omninew_2025, xai_grok_imagine_2026, wu2026consid}. In professional production workflows, however, a prompt-generated video rarely satisfies the desired result in a single pass; it typically undergoes multiple rounds of targeted refinement, such as moving objects, adjusting camera motion, or adding visual effects, before it can be used. As a result, instruction-guided video editing \citep{jiang2025vace, wei2025univideo, wan2025wan}, where a user specifies a natural-language instruction to modify an existing video, has therefore become an essential component of AI-assisted filmmaking.

Despite rapid progress, the evaluation of video editing and visual effects (VFX) remains fundamentally unresolved. Unlike generic video generation, video editing must answer at least three distinct questions: Did the model execute the requested edit? Is the edited video visually coherent and temporally plausible? Did it preserve content that should have remained unchanged? These requirements expose two major bottlenecks. First, the field lacks large-scale human-annotated resources that contain complete editing triplets—the source video, the editing instruction, and the edited result—along with fine-grained quality labels. Second, evaluation still relies heavily on costly manual inspection or generic vision-language model (VLM) judges that are not designed for video-editing-specific assessment. The absence of a dedicated automatic evaluator makes both systematic benchmarking and preference-based optimization difficult.

Existing resources address only parts of this problem. Benchmarks like EditBoard~\citep{chen2025editboard},  FiVE-Bench~\citep{li2025five}, and IVE-Bench~\citep{chen2025ivebench} provide instructions without edited outputs; OpenVE ~\citep{he2025openve} offers scale but relies heavily on automated generation and filtering rather than human annotation; VE-Bench~\citep{sun2024bench} included edited videos and human scores but reduced quality to a single scalar and are built on older editing systems. On the reward-model side, prior work focuses on image editing or video generation quality rather than video editing itself \citep{wu2025editreward, liu2025improving}. As a result, there is a pressing need for a benchmark and evaluator that jointly capture instruction faithfulness, rendering quality, and preservation of unedited content.

To address these gaps, we introduce \GD, \GR, and \GB. \GD contains 5,049 human-annotated video editing examples spanning 9 major categories and 32 fine-grained subcategories. Each example contains a source video, an editing instruction, and an edited result produced by a diverse mixture of commercial systems, open-source models, and agentic editing pipelines. Trained annotators score each example along three decoupled dimensions: Instruction Following (IF), Rendering Quality (RQ), and Edit Exclusivity (EE). This design is central to the benchmark: an edit may be semantically wrong but visually clean, or visually strong while unnecessarily modifying non-target content. Building on \GD, we train \GR, a reward model that takes the source video, the editing instruction, and the edited video as input and predicts per-dimension quality scores via ordinal regression. We further release \GB, a standardized benchmark of 300 curated video-prompt pairs for systematic model comparison, and use \GR to evaluate representative commercial and open-source editing systems under the same multi-dimensional protocol. Our contributions are summarized as follows:
\begin{itemize}[leftmargin=*, itemsep=1pt, topsep=2pt]
    \item We construct \GD, a human-annotated dataset of 5,049 video editing examples across 9 main categories and 32 subcategories, generated by a diverse mixture of commercial, open-source, and agentic editing systems. Each example is scored on a 4-point rubric along three decoupled dimensions: IF, RQ, and EE. We further release \GB, a standardized benchmark of 300 curated video-prompt pairs for comparing editing systems.

    \item We propose \GR, the first dedicated reward model for video editing quality assessment. \GR jointly reasons over the source video, the editing instruction, and the edited result, and predicts multi-dimensional quality scores with an ordinal regression objective.

    \item We conduct comprehensive experiments showing that \GR aligns more strongly with human judgments than generic VLM judges and prior reward-model baselines on both standard IQA/VQA metrics and group-wise preference evaluation. We further apply \GR to benchmark representative commercial and open-source video editing systems, exposing task-dependent strengths and persistent weaknesses in instruction following and edit locality.
\end{itemize}

\section{Related Work}
\label{sec:related_work}

\subsection{Instruction-Guided Video Editing}
\label{sec:rw_editing}

Instruction-guided video editing aims to modify a video according to natural-language instructions while preserving unrelated content~\cite{jiang2025vace, wei2025univideo, li2026physics, motamed2026void, burgert2025motionv2v, gao2026pisco}. Early methods extended image editing pipelines to the temporal domain, typically by introducing temporal attention or consistency modules on top of text-to-image diffusion models \citep{guo2023animatediff, yang2023diffusion}. More recent approaches adopt video-native diffusion or flow-matching architectures. Representative research models include VACE \citep{jiang2025vace}, UniVideo \citep{wei2025univideo}, and the broader Wan family \citep{wan2025wan, cheng2025wan}. Alongside them, commercial systems such as Kling Omni, Grok Imagine, Luma Ray2, and the commercial Wan 2.6 service variant have reached practical quality levels \citep{klingai_omninew_2025, xai_grok_imagine_2026, lumaray2_2025, cheng2025wan}. The resulting ecosystem is highly heterogeneous, with different systems excelling on different editing types, which makes standardized evaluation increasingly important.

\subsection{Video Editing Quality Evaluation}
\label{sec:rw_evaluation}

Evaluating video editing quality is intrinsically multi-faceted. Conventional metrics such as CLIP score, SSIM, and LPIPS capture only narrow aspects of the problem and do not directly measure instruction fidelity, temporal consistency, or unintended edits \citep{radford2021learning, zhang2018unreasonable}. VBench and VBench++ provide broad evaluation suites for video generation, but they are not designed for editing, where the source and edited videos must be considered jointly \citep{huang2024vbench, huang2025vbench++}. Several editing-oriented resources have been introduced more recently. \texttt{EditBoard} \citep{chen2025editboard} and \texttt{FiVE} \citep{li2025five} provide useful task-oriented protocols, but at limited scope or scale. \texttt{OpenVE-3M} \citep{he2025openve} provides scale without human quality annotation. \texttt{IVE-Bench} \citep{chen2025ivebench} includes source videos and instructions with a multi-dimensional protocol, but no edited results. \texttt{VE-Bench} \citep{sun2024bench} includes edited videos and human scores, but reduces quality to a single scalar MOS. In contrast, \GD provides large-scale human-annotated video editing examples with decoupled quality labels tailored specifically to the editing setting.

\subsection{Reward Models for Visual Generation}
\label{sec:rw_alignment}

The success of RLHF in language modeling has motivated analogous efforts in visual generation. For image generation, reward models such as ImageReward, HPS, and PickScore learn to approximate human preference signals from large-scale annotations \citep{xu2023imagereward, wu2023human, kirstain2023pick}. This line of work has extended to image editing: EditReward trains a multi-dimensional reward model for instruction-guided image editing and demonstrates value for both evaluation and data curation \citep{wu2025editreward}. In the video domain, VideoScore, VideoReward, DenseDPO, WorldScore, and Pulse model human preferences or preference-driven alignment primarily for video generation \citep{he2024videoscore, liu2025improving, wu2025densedpo, duan2025worldscore, gao2026pulse}. VE-Bench also includes a video editing assessor, but it predicts only a single scalar score and is tied to an earlier benchmark setting \citep{sun2024bench}. These methods do not explicitly reason over the relationship between the source video and the edited result. \GR addresses this gap by jointly processing the original video, the editing instruction, and the edited result, and by predicting multi-dimensional quality scores tailored to video editing.

\section{\GD and \GB}
\label{sec:benchmark}

We present \GD, a human-annotated dataset for video editing quality evaluation, and \GB, a standardized benchmark for systematic model comparison. \GD contains 5,049 editing examples---4,200 for training and 849 for testing---covering 9 major categories and 32 subcategories, each annotated along three decoupled quality dimensions: Instruction Following (IF), Rendering Quality (RQ), and Edit Exclusivity (EE). \GB contains 300 curated (raw video, editing prompt) pairs for evaluating and comparing video editing models under a standardized protocol. This section describes the data collection process, annotation protocol, reliability check, and key dataset statistics.

\Cref{tab:dataset_comparison} compares \GD with existing video editing datasets along three properties that are particularly important for reward modeling: whether the dataset includes edited outputs, whether the scores come from human annotation, and whether quality is decomposed into multiple dimensions. These properties matter because reward-model training requires actual edited results, reliable human supervision, and labels that distinguish different failure modes. Several recent resources provide prompts without edited outputs \citep{chen2025editboard, li2025five, chen2025ivebench}; others rely on automated filtering or judge models rather than trained annotators \citep{li2025five, he2025openve}; and some collapse editing quality into a single scalar. \GD is the only dataset in this comparison that satisfies all three conditions simultaneously.

\begin{table*}[h]
\centering
\caption{Comparison of \GD with existing video editing datasets. In the ``\#Cate'' column, entries such as ``8/35'' denote 8 major categories and 35 subcategories. ``Human Ann.'' indicates whether quality scores are provided by human annotators. ``Multi-Dim.'' indicates whether the evaluation is decomposed into multiple quality dimensions. ``Editing Systems'' summarizes the diversity of models used to generate edited videos.}
\label{tab:dataset_comparison}
\tablestyle{5pt}{1.15}
\resizebox{\textwidth}{!}{
\begin{tabular}{l c c c c c c l}
\shline
Dataset & \#Videos & \#Pairs & \#Cate & Edited Videos & Human Ann. & Multi-Dim. & Editing Systems \\
\shline
\texttt{VE-Bench}~\citep{sun2024bench} & 169 & 1,170 & 6 & \CK & \CK & \lX & 8 SD-based open-source (2024) \\
\texttt{EditBoard}~\citep{chen2025editboard} & \ldash & \ldash & 4 & \lX & \ldash & \ldash & \ldash \\
\texttt{FiVE}~\citep{li2025five} & $\sim$100 & 420 & 6 & \lX & \ldash & \ldash & \ldash \\
\texttt{OpenVE-3M}~\citep{he2025openve} & 1M & 3M & 8 & \CK & \lX & \CK & Open-source + agentic (2025) \\
\texttt{IVE-Bench}~\citep{chen2025ivebench} & 600 & \ldash & 8/35 & \lX & \ldash & \ldash & \ldash \\
\rowcolor{blue!8}
\GD (Ours) & 1,988 & 5,049 & 9/32 & \CK & \CK & \CK & 4 Commercial + Open + Agentic (2026) \\
\shline
\end{tabular}
}
\end{table*}

\subsection{Data Collection}
\label{sec:data_collection}

\paragraph{Source videos.} We curate source videos from open-source video datasets including Open-Sora~\citep{zheng2024opensora} and OpenVid-1M~\citep{nan2024openvid}, supplemented with privately collected footage for additional diversity. We filter the initial pool for quality and usability, including sufficient resolution, duration, and temporal continuity, remove NSFW content, and then sample across scene categories and content types. The final set contains 1,419 source videos spanning 10 scene categories, summarized in \Cref{fig:data_statistics}(c).

\begin{wrapfigure}{r}{0.33\textwidth}
\centering
\vspace{-5mm}
\includegraphics[width=\linewidth]{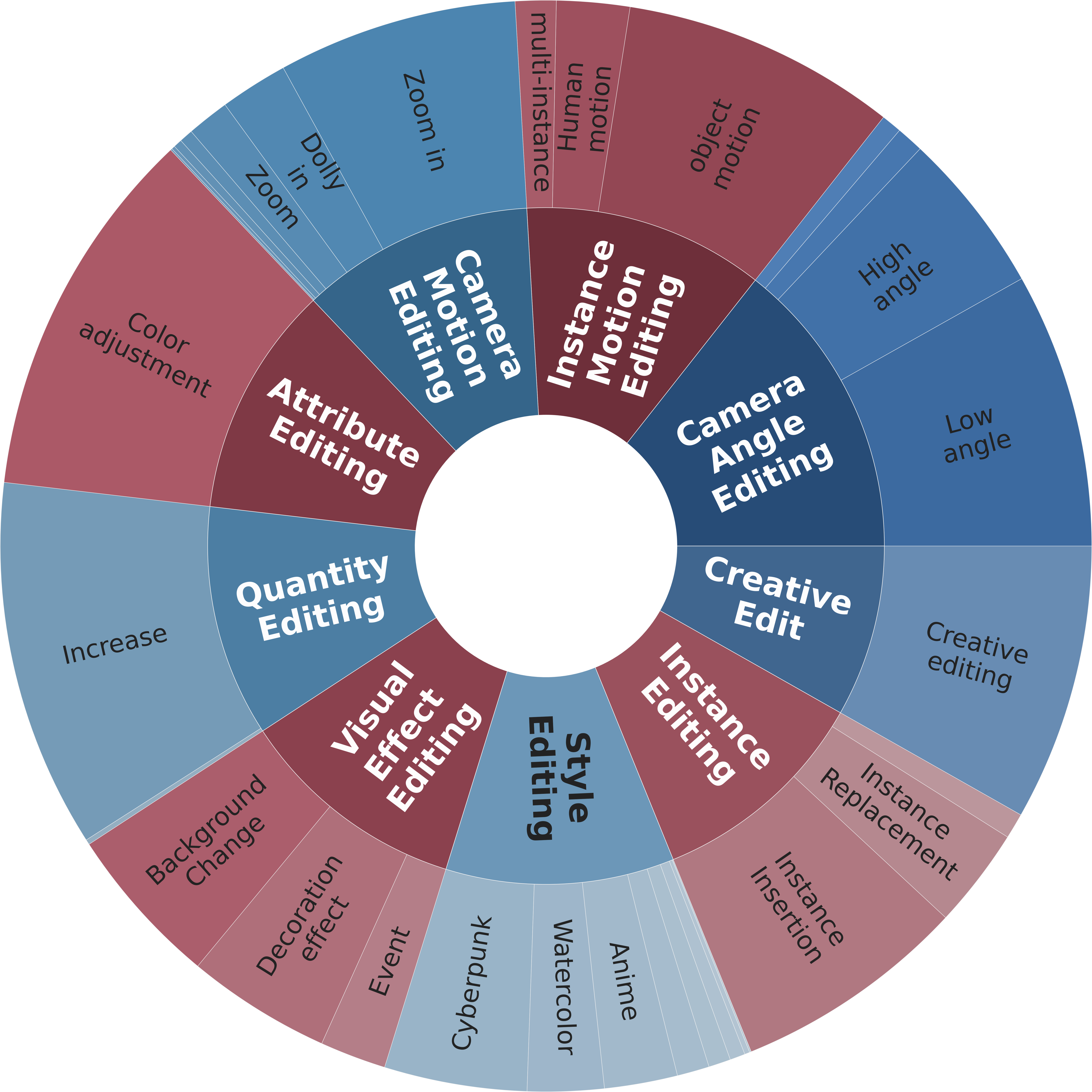}
\caption{Task hierarchy of the 9 main editing categories and 32 subcategories in \GD.}
\label{fig:task_hierarchy}
\vspace{-8mm}
\end{wrapfigure}

\paragraph{Editing instructions.} We design instructions to cover 9 major editing categories and 32 subcategories, illustrated in \Cref{fig:task_hierarchy}. To improve task-video compatibility, we use Gemini 3 Flash~\citep{deepmind_gemini_flash_2025} to analyze video content, assign suitable editing categories, and generate matched prompts. Low-confidence assignments are discarded. This process yields broad task coverage while keeping the editing instructions grounded in the source content.

\paragraph{Edited video generation.} For each (source video, instruction) pair, we collect edited videos from a diverse mixture of commercial systems, open-source models, and agentic editing pipelines. This diversity is important because it exposes the benchmark to a broad range of quality levels and failure modes rather than the behavior of a single model family. Detailed model lists and pipeline descriptions are provided in \Cref{sec:appendix_pipelines}.

\subsection{Annotation Protocol}
\label{sec:annotation}

Each editing example is evaluated on a 4-point scale along three decoupled dimensions.

\paragraph{Instruction Following (IF).} IF measures whether the edit satisfies the semantic requirements of the instruction. A score of 4 indicates that all requested edits are completed correctly, while a score of 1 indicates failure, contradiction, or an edit that is largely unrelated to the instruction.

\paragraph{Rendering Quality (RQ).} RQ evaluates visual quality, including clarity, naturalness, temporal stability, and the absence of artifacts such as flickering, ghosting, blur, or distortion. This dimension is scored independently of whether the instruction is followed.

\paragraph{Edit Exclusivity (EE).} EE assesses whether the model changes only the intended target region without introducing unnecessary modifications elsewhere. In our annotation guide, score 4 means that no clear non-target change is introduced, score 3 corresponds to one localized non-target change, score 2 corresponds to two to three clear non-target changes or one large unintended background change, and score 1 indicates widespread or global over-editing.

A summary of the rubric is provided in \Cref{tab:rubric}, with the complete annotation guide in \Cref{sec:appendix_guide}. The key principle of the protocol is that IF, RQ, and EE are scored independently. For example, if the instruction is ``turn the apple into a banana'' but the model returns the unchanged video with excellent visual quality, the correct labels are IF = 1, RQ = 4, and EE = 4. This decoupling prevents semantic success, visual fidelity, and locality preservation from contaminating one another. All annotators complete a calibration phase with detailed guidelines and reference examples before annotation begins.

\begin{table}[t]
\centering
\caption{Summary of the 4-point scoring rubric for each annotation dimension.}
\label{tab:rubric}
\tablestyle{3pt}{1.1}
\setlength{\tabcolsep}{9pt}
\resizebox{\columnwidth}{!}{
\begin{tabular}{l p{95pt} p{115pt} p{120pt} p{80pt}}
\shline
 & Score 4 & Score 3 & Score 2 & Score 1 \\
\shline
IF & All requested edits completed correctly & Core edit completed with minor deviation & Partial execution with major semantic deviation & Failure, contradiction, or unrelated edit \\
\hline
RQ & Clear, stable, and artifact-free & Minor but noticeable degradation & Clear quality failure with recurrent artifacts & Severe visual breakdown \\
\hline
EE & No clear non-target change & One clear non-target change & Two--three non-target changes or one large unintended change & Global or widespread over-editing \\
\shline
\end{tabular}
}
\vspace{-0.3cm}
\end{table}

\subsection{Annotation Reliability}
\label{sec:quality_verification}

To assess annotation reliability, we conduct a targeted cross-check study. We randomly sample 550 examples from the dataset and re-annotate them with a double-annotation strategy using a new group of annotators independent of the original raters. Since the labels lie on a four-point ordinal scale, we report two direct agreement measures: exact agreement and within-1-point agreement.

\begin{wraptable}{r}{0.5\textwidth}
\centering
\vspace{-5mm}
\caption{Inter-annotator agreement on the 550-sample cross-check subset. Higher values indicate stronger agreement.}
\label{tab:iaa}
\tablestyle{6pt}{1.15}
\resizebox{0.5\textwidth}{!}{
\begin{tabular}{l c c c}
\shline
Metric & IF & RQ & EE \\
\shline
Exact Agreement (\%) & 75.2 & 87.2 & 72.2 \\
Within-1 Agreement (\%) & 93.5 & 97.2 & 91.7 \\
\shline
\end{tabular}}
\vspace{-3mm}
\end{wraptable}

\Cref{tab:iaa} shows strong agreement under this cross-check protocol. Within-1 agreement exceeds 91\% on all three dimensions, reaching 93.5\% for IF, 97.2\% for RQ, and 91.7\% for EE, while exact agreement remains high at 75.2\%, 87.2\%, and 72.2\%, respectively. This pattern is intuitive: rendering quality is the easiest dimension to align on, while instruction following and edit exclusivity involve more borderline cases around partial success and acceptable non-target change. Although limited in scale, this study provides a useful sanity check that the three-dimensional labels are stable enough for training and evaluation. Additional details are provided in \Cref{sec:appendix_iaa}.

\subsection{Dataset Statistics and Analysis}
\label{sec:bench_analysis}

We present several analyses of \GD to characterize the dataset and motivate its three-dimensional design. Extended analysis is provided in \Cref{sec:appendix_analysis}.

\begin{figure*}[t]
\centering
\includegraphics[width=\textwidth]{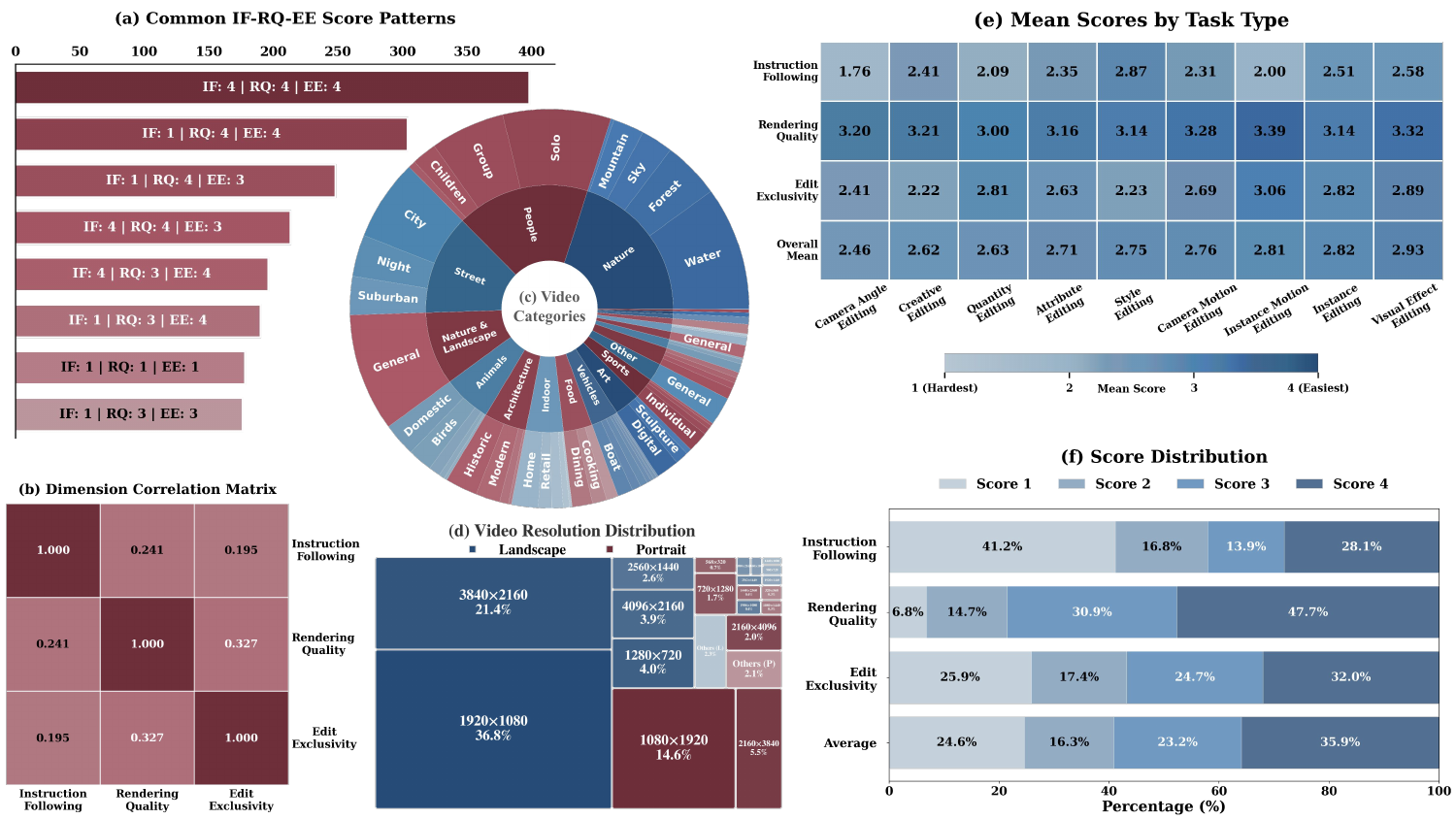}
\caption{Overview of dataset statistics for \GD. Panel (a) shows common IF--RQ--EE score patterns; (b) reports pairwise dimension correlations; (c) summarizes video-category coverage; (d) shows the video-resolution distribution; (e) reports mean scores by task type; and (f) shows score distributions across annotation dimensions. Together they show that \GD spans diverse content and resolutions, exhibits heterogeneous task difficulty, and captures clear variation in difficulty across editing tasks.}
\label{fig:data_statistics}
\vspace{-0.3cm}
\end{figure*}

\paragraph{Score distributions and score patterns.} \Cref{fig:data_statistics} shows that the three quality dimensions follow clearly different distributions, which justifies decoupled evaluation. IF is the most polarized: 41.2\% of samples receive score 1, while 28.1\% receive score 4, indicating that many edits either fail outright or satisfy the instruction well. RQ is much more right-skewed, with 78.6\% of samples receiving scores 3 or 4 and only 6.8\% receiving score 1, suggesting that visual plausibility is often easier to achieve than semantic correctness. EE is more balanced across score levels. The common score-pattern panel reinforces this point: while $(4,4,4)$ is the most frequent triplet, several of the next most common patterns are cases such as $(1,4,4)$ and $(1,4,3)$, where the output looks plausible but does not follow the instruction.

\paragraph{Task difficulty across editing types.} The task-type heatmap reveals substantial variation in difficulty across editing categories. Camera Angle Editing is the hardest category overall, with IF = 1.76 and an overall mean of 2.46, while Quantity Editing is also challenging on IF at 2.09. By contrast, Style Editing reaches the highest IF at 2.87, and Visual Effect Editing attains the highest overall mean at 2.93. RQ remains relatively stable across task types, between 3.00 and 3.39, again suggesting that current systems are better at producing visually plausible outputs than at satisfying complex instructions. EE varies more strongly, with Instance Motion Editing reaching 3.06 while Creative Editing and Style Editing are lower at 2.22 and 2.23.

\paragraph{Coverage and dimension independence.} Panels (c) and (d) show that \GD spans diverse scene types and video formats rather than collapsing to a single narrow distribution. Large groups such as Nature, People, and Street are well represented, while finer-grained content remains present in the long tail. The resolution distribution is also broad: 1920$\times$1080 is the largest bucket at 36.8\%, 3840$\times$2160 contributes 21.4\%, and portrait videos such as 1080$\times$1920 remain substantial at 14.6\%. This coverage matters because editing difficulty depends on both semantic content and visual format. To verify that IF, RQ, and EE capture distinct aspects of editing quality, we further compute pairwise correlations over the full dataset. All correlations remain weak: IF--RQ is 0.241, IF--EE is 0.195, and RQ--EE is 0.327. These low correlations support the three-axis annotation design and indicate that a single scalar score would obscure important failure modes.

\section{\GR: Human-Aligned Video Editing Reward Model}
\label{sec:method}

We present the design and training of \GR, a reward model that predicts human-aligned quality scores for video editing results. Unlike existing reward models that target either image editing or video generation, \GR is specifically designed for the video editing setting, where quality assessment must jointly consider the original video, the editing instruction, and the edited output.

\subsection{Problem Formulation}
\label{sec:problem}

Given an original video $V_o = \{v_o^t\}_{t=1}^{T}$, an editing instruction $P$, and the corresponding edited video $V_e = \{v_e^t\}_{t=1}^{T}$, our goal is to predict quality scores that align with human judgment along the three annotation dimensions defined in \Cref{sec:annotation}:
\begin{equation}
[s_{\text{IF}}, s_{\text{RQ}}, s_{\text{EE}}] = \mathcal{F}(V_o, P, V_e),
\end{equation}
where each score lies on the ordinal scale $\{1,2,3,4\}$.

The three dimensions require different reasoning. Instruction Following evaluates the semantic execution of the requested edit. Rendering Quality measures visual fidelity and temporal consistency. Edit Exclusivity compares the original and edited videos to detect unintended modifications outside the target region. A single holistic score would obscure these distinct failure modes, which motivates the multi-dimensional formulation of \GR.

\subsection{Architecture}
\label{sec:architecture}

\GR is instantiated on the Qwen3-VL-Instruct family~\citep{yang2025qwen3} at two scales, 4B and 32B, which correspond to \GRs and \GRl in the experiments. In both variants, the model jointly processes the original video, the edited video, and the editing instruction. This design allows the backbone to compare the edited result against both the requested change and the source content, which is essential for assessing semantic faithfulness, rendering quality, and unintended edits within one shared representation.

We introduce three learnable special tokens, \texttt{<|IF\_reward|>}, \texttt{<|RQ\_reward|>}, and \texttt{<|EE\_reward|>}, to query the three target dimensions. Their final hidden states are passed to a shared reward head, which produces the ordinal logits used for prediction. This token-based design gives each dimension its own query while preserving a single backbone for joint multimodal reasoning.

\subsection{Ordinal Regression Objective}
\label{sec:training}

The bimodal distribution of IF scores (\Cref{sec:bench_analysis}) and the ordinal nature of the 4-point scale motivate the use of ordinal regression rather than standard L2 loss. We adopt ordinal regression~\citep{shaham2020deep}, which models the score as a sequence of ordered threshold decisions instead of an unconstrained scalar regression target.

For each dimension, the reward head predicts three ordered probabilities corresponding to whether the score is greater than 1, 2, and 3. Training applies binary cross-entropy to these ordered threshold predictions under the formulation:
\begin{equation}
\mathcal{L} = \sum_{d} \frac{1}{K-1} \sum_{k=1}^{K-1} \text{BCE}\Big(\sigma(z_d^k),\; \mathbf{1}[y_d > k] \;\Big|\; y_d \geq k\Big),
\label{eq:corn_loss}
\end{equation}
where the conditional constraint $y_d \geq k$ ensures that each threshold is trained only on relevant samples, preserving the ordinal structure.

At inference, we convert these ordered probabilities into a continuous score on $[1,4]$ by taking their expected value:
\begin{equation}
\hat{s}_d = 1 + \sum_{k=1}^{K-1} P(Y > k).
\end{equation}
This soft prediction is used in all reported experiments.

\subsection{Training Details}
\label{sec:training_details}

\paragraph{Data and video processing.} We train \GR on the 4,200-example training split of \GD and evaluate on the 849-example test split, with the split stratified across editing categories and pipelines. For each example, we uniformly sample both the original and edited videos at 4 FPS and cap the frame resolution at 399,360 pixels, approximately 632 $\times$ 632, while preserving native aspect ratios through Qwen3-VL's dynamic-resolution mechanism. The two videos are sampled with aligned temporal indices to support direct comparison, and the maximum sequence length is set to 32,768 tokens.

\paragraph{Optimization.} We use a two-stage training schedule. In the first stage, lasting 1 epoch, we freeze all pretrained parameters and train only the newly introduced reward tokens and reward head. In the second stage, lasting 49 epochs, we unfreeze and fine-tune the language backbone and visual-language merger together with the reward head and reward tokens, while keeping the vision tower frozen. We optimize with AdamW using learning rates of $1 \times 10^{-5}$ for the language-side parameters and $5 \times 10^{-5}$ for the reward tokens, cosine decay, and a 15\% warmup ratio. Training is performed in bf16 on 8 GPUs with an effective batch size of 8, and all three reward dimensions are optimized jointly with equal loss weights.

\section{Experiments}
\label{sec:experiments}

We conduct comprehensive experiments to evaluate \GR as a video editing quality assessor. We compare against generic VLM-as-judge baselines and prior reward models, analyze global agreement with standard IQA/VQA metrics, and further test whether the learned scores preserve local human preferences within directly comparable candidate sets.

\subsection{Experimental Setup}
\label{sec:exp_setup}

\paragraph{Evaluation metrics.} Our primary evaluation follows standard IQA/VQA protocol. We report Spearman Rank-Order Correlation Coefficient (SRCC), Kendall Rank-Order Correlation Coefficient (KRCC), Pearson Linear Correlation Coefficient (PLCC), and Root Mean Squared Error (RMSE) in \Cref{sec:classical_metrics}. SRCC and KRCC are computed on raw predictions, while PLCC and RMSE are computed after the standard four-parameter logistic calibration. We complement these global correlation metrics with a group-wise preference metric, Pairwise Accuracy, in \Cref{sec:ranking_results}. Detailed metric definitions and the calibration protocol are provided in \Cref{sec:appendix_classical_metrics}.

\paragraph{Baselines.} We compare \GR against three types of baselines:
\begin{itemize}[leftmargin=*, itemsep=1pt, topsep=2pt]
    \item \textit{VLM-as-a-Judge}: Qwen3.5-397B, Qwen3.5-122B~\citep{yang2025qwen3}, Gemini-3.1-Pro, Gemini-3.1-Flash-Lite, Gemini-2.5-Flash~\citep{deepmind_gemini_pro_2025, deepmind_gemini_flash_2025}, and Seed-2.0-Lite, Seed-1.6~\citep{bytedance2025seed16flash}. Each model receives the source video, editing instruction, and edited video, and is prompted to score editing quality on the same 1--4 rubric used in human annotation.
    \item \textit{EditReward}: an image editing reward model with two output heads, one aligned with instruction following and one aligned with generic visual quality~\citep{wu2025editreward}.
    \item \textit{VE-Bench}: a video editing reward model that predicts a single scalar quality score~\citep{sun2024bench}.
\end{itemize}

\paragraph{Implementation details.} We instantiate \GR at two scales, \GRs and \GRl, using Qwen3-VL backbones at 4B and 32B with the same architecture and training objective. Both models are trained on the 4,200-example training split and evaluated on the 849-example test split. We sample both the original and edited videos at 4 FPS with a maximum frame resolution of 399,360 pixels while preserving aspect ratio, and train in bf16 with an effective batch size of 8. For VLM-as-judge baselines, we use a shared rubric-aligned prompt over the same source-video/instruction/edited-video triplet. In our evaluation, the human overall score is defined as the arithmetic mean of IF, RQ, and EE. For \GR and VLM-as-judge baselines, the overall prediction is the mean of the three predicted dimension scores; for EditReward, it is the mean of its two native heads; and for VE-Bench, it is the model's native scalar output. Additional implementation details are provided in \Cref{sec:appendix_experiment_details}.

\subsection{Results on Standard IQA/VQA Metrics}
\label{sec:classical_metrics}

Following standard IQA/VQA practice, we evaluate all methods with SRCC, KRCC, PLCC, and RMSE. The Overall columns in \Cref{tab:classical_metrics} report agreement on the human overall score rather than a separate learned target.

\begin{table*}[h]
\centering
\caption{Results on standard IQA/VQA metrics. SRCC, KRCC, and PLCC are higher-is-better; RMSE is lower-is-better. PLCC and RMSE are computed after logistic calibration. Overall denotes correlation on the human overall score, defined as the mean of IF, RQ, and EE. For \GR and VLM-as-judge baselines, the overall prediction is the mean of the three predicted dimension scores; EditReward uses the mean of its two native heads, and VE-Bench uses its native scalar overall score.}
\label{tab:classical_metrics}
\tablestyle{3.6pt}{1.3}
\setlength{\tabcolsep}{2pt}
\resizebox{\textwidth}{!}{
\begin{tabular}{l cccc cccc cccc cccc}
\shline
\multirow{2}{*}{Method} & \multicolumn{4}{c}{SRCC$\uparrow$} & \multicolumn{4}{c}{KRCC$\uparrow$} & \multicolumn{4}{c}{PLCC$\uparrow$} & \multicolumn{4}{c}{RMSE$\downarrow$} \\
\cmidrule(lr){2-5} \cmidrule(lr){6-9} \cmidrule(lr){10-13} \cmidrule(lr){14-17}
 & IF & RQ & EE & Overall & IF & RQ & EE & Overall & IF & RQ & EE & Overall & IF & RQ & EE & Overall \\
\hline
\multicolumn{17}{c}{\textit{VLM-as-a-Judge}} \\
\shline
Seed-1.6 & 0.686 & 0.618 & 0.504 & 0.630 & 0.605 & 0.573 & 0.447 & 0.508 & 0.684 & 0.608 & 0.591 & 0.701 & 0.918 & 0.798 & 0.917 & 0.565 \\
Seed-2.0-Lite & 0.545 & 0.544 & 0.697 & 0.720 & 0.483 & 0.497 & 0.527 & 0.607 & 0.616 & 0.594 & 0.729 & 0.768 & 0.984 & 0.815 & 0.791 & 0.510 \\
Qwen3.5-122B & 0.379 & 0.563 & 0.658 & 0.631 & 0.327 & 0.523 & 0.573 & 0.520 & 0.378 & 0.663 & 0.601 & 0.685 & 1.165 & 0.752 & 0.893 & 0.578 \\
Qwen3.5-397B & 0.572 & 0.422 & 0.654 & 0.601 & 0.506 & 0.384 & 0.587 & 0.497 & 0.615 & 0.624 & 0.692 & 0.657 & 0.992 & 0.785 & 0.820 & 0.598 \\
Gemini-3.1-Pro & 0.731 & 0.518 & 0.681 & 0.752 & 0.559 & 0.459 & 0.584 & 0.608 & \textbf{0.754} & 0.510 & 0.644 & 0.726 & 0.826 & 0.864 & 0.788 & 0.546 \\
Gemini-3.1-Flash-Lite & 0.309 & 0.302 & 0.661 & 0.574 & 0.283 & 0.277 & 0.555 & 0.436 & 0.316 & 0.425 & 0.673 & 0.505 & 1.194 & 0.910 & 0.840 & 0.685 \\
Gemini-2.5-Flash & 0.256 & 0.217 & 0.581 & 0.383 & 0.236 & 0.195 & 0.544 & 0.296 & 0.256 & 0.491 & 0.569 & 0.478 & 1.216 & 0.875 & 0.934 & 0.697 \\
\hline
\multicolumn{17}{c}{\textit{Previous Reward Models}} \\
\hline
EditReward~\citep{wu2025editreward} & 0.453 & -0.211 & \ldash & 0.558 & 0.342 & -0.164 & \ldash & 0.411 & 0.455 & 0.317 & \ldash & 0.580 & 1.113 & 0.844 & \ldash & 0.631 \\
VE-Bench~\citep{sun2024bench} & \ldash & \ldash & \ldash & 0.214 & \ldash & \ldash & \ldash & 0.150 & \ldash & \ldash & \ldash & 0.238 & \ldash & \ldash & \ldash & 0.752 \\
\hline
\multicolumn{17}{c}{\textit{Ours}} \\
\hline
\rowcolor{green!10}
\GRs & 0.714 & \textbf{0.690} & 0.693 & 0.760 & 0.564 & \textbf{0.574} & 0.556 & 0.595 & 0.704 & \textbf{0.793} & 0.710 & 0.771 & 0.888 & \textbf{0.642} & 0.764 & 0.493 \\
\rowcolor{green!10}
\GRl & \textbf{0.754} & 0.681 & \textbf{0.717} & \textbf{0.780} & \textbf{0.612} & 0.567 & \textbf{0.597} & \textbf{0.616} & 0.751 & 0.792 & \textbf{0.732} & \textbf{0.790} & \textbf{0.825} & 0.643 & \textbf{0.740} & \textbf{0.475} \\
\shline
\end{tabular}
}
\vspace{-0.2cm}
\end{table*}

\paragraph{Overall results.} Both \GR variants clearly outperform prior reward-model baselines on the human overall score. \GRl is strongest overall, achieving 0.780 SRCC, 0.616 KRCC, 0.790 PLCC, and 0.475 RMSE, while \GRs follows closely at 0.760, 0.595, 0.771, and 0.493. The margin over prior reward models is substantial: EditReward reaches 0.558 overall SRCC and 0.631 RMSE, whereas VE-Bench drops further to 0.214 SRCC and 0.752 RMSE.

\paragraph{Dimension-wise behavior.} The two \GR scales show complementary strengths. \GRl is best on IF and EE, with the strongest rank correlation and calibration on both dimensions. \GRs is slightly stronger on RQ across all four standard metrics, which suggests that larger scale mainly helps instruction faithfulness and edit exclusivity, while rendering-quality prediction is already close to saturation at 4B scale. This is consistent with the dataset statistics in \Cref{sec:bench_analysis}: RQ is both less ambiguous and more concentrated than IF.

\paragraph{Baseline comparison.} Strong VLM judges remain competitive on a few individual columns, but they do not match the consistency of \GR across dimensions and metrics. More importantly, the gap to previous reward models is large and systematic. EditReward remains somewhat useful on IF, but its negative RQ correlations indicate a clear mismatch between image-editing supervision and video-editing assessment; it also has no dedicated EE head. VE-Bench predicts only a single scalar score and therefore cannot support per-dimension analysis, while even its overall agreement remains weak. These results support the need for a reward model that jointly reasons over the source video, the editing instruction, and the edited output.

\begin{figure}[t]
\centering
\includegraphics[width=\columnwidth]{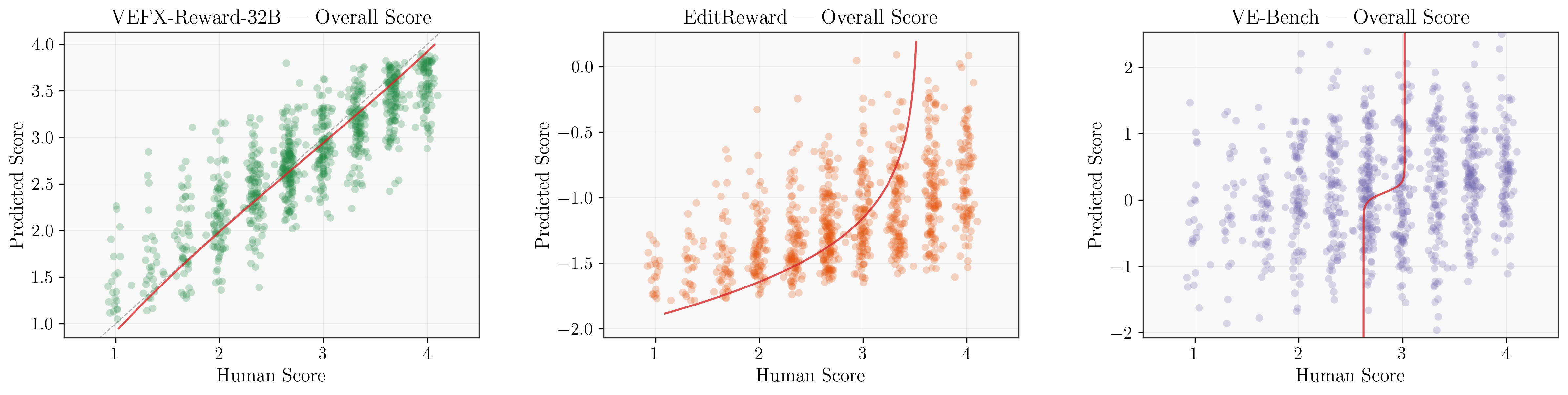}
\caption{Predicted overall scores versus human overall scores for \GRl, EditReward, and VE-Bench. Here the human overall score is defined as the mean of IF, RQ, and EE. \GRl exhibits a tight monotonic trend that closely follows the human score axis, whereas EditReward shows a weaker and more nonlinear relationship, and VE-Bench displays substantially larger dispersion with limited sensitivity to score differences. Additional scatter plots are provided in \Cref{sec:appendix_classical_metrics}.}
\label{fig:classical_scatter_avg}
\vspace{-0.2cm}
\end{figure}

\paragraph{Scatter-plot analysis.} \Cref{fig:classical_scatter_avg} provides a qualitative comparison of the three reward models on the human overall score. \GRl shows a clear monotonic increase and a relatively tight concentration around the fitted trend, indicating that its predictions preserve both ordering and score magnitude more faithfully. EditReward still captures a coarse positive trend, but the response is more nonlinear and compressed. VE-Bench exhibits the weakest alignment, with much larger dispersion at nearly every human score level. This visual evidence is fully consistent with the quantitative results in \Cref{tab:classical_metrics}.

\subsection{Group-wise Preference Evaluation}
\label{sec:ranking_results}

Standard IQA/VQA metrics measure global correlation with human scores, but reward models are often used in a more local setting: given several candidate edits for the same source video and instruction, the model should prefer the better one. We therefore add a group-wise preference evaluation that measures whether a reward model preserves human ordering within directly comparable candidate sets.

\paragraph{Pairwise Accuracy.} Each ranking group $g$ contains all candidate edits that share the same raw video and editing instruction; candidate edits may come from different editing systems, but comparisons are performed only within the group and never across groups. We enumerate all candidate pairs in each group and compare the predicted ordering with the ground-truth ordering. If the ground truth is tied, the pair is counted as correct regardless of the prediction; if the prediction is tied but the ground truth is not, it receives a score of $0.5$. The dataset-level Pairwise Accuracy is
\begin{equation}
\mathrm{PairAcc} = \frac{\sum_{g=1}^{G} \sum_{(i,j)\in \mathcal{P}_g} \mathrm{Acc}_{ij}}{\sum_{g=1}^{G} |\mathcal{P}_g|},
\end{equation}
where $\mathrm{Acc}_{ij}=1$ if the predicted order matches the human order. Unlike the global IQA/VQA metrics above, Pairwise Accuracy depends only on relative ordering within each candidate group and is therefore insensitive to score-scale mismatch across models.

\begin{table}[h]
\centering
\caption{Group-wise preference evaluation using Pairwise Accuracy. Overall denotes performance on the human overall score, defined as the mean of IF, RQ, and EE.}
\label{tab:ranking_results}
\tablestyle{5pt}{1.1}
\setlength{\tabcolsep}{16pt}
\resizebox{0.7\columnwidth}{!}{
\begin{tabular}{l cccc}
\shline
Model & IF & RQ & EE & Overall \\
\shline
EditReward~\citep{wu2025editreward} & 0.8283 & 0.5629 & 0.5317 & 0.7919 \\
VE-Bench~\citep{sun2024bench} & 0.7351 & 0.8127 & 0.7143 & 0.6651 \\
\rowcolor{green!10}
\GRs & 0.9120 & \textbf{0.9309} & 0.9167 & 0.8628 \\
\rowcolor{green!10}
\GRl & \textbf{0.9366} & 0.9111 & \textbf{0.9196} & \textbf{0.8723} \\
\shline
\end{tabular}
}
\vspace{-0.25cm}
\end{table}

\paragraph{Preference results.} Both \GR variants substantially outperform previous reward models on group-wise preference consistency. \GRl achieves the best overall Pairwise Accuracy at 0.872, while \GRs remains close at 0.863, indicating that most relative preference signal is already captured at 4B scale. EditReward remains somewhat competitive on IF because one of its heads is naturally aligned with instruction following, but it performs poorly on RQ and EE because it is an image editing reward model without video-native temporal reasoning or a dedicated EE concept. VE-Bench shows moderate ordering ability, but its single-score design limits fine-grained candidate comparison. Together, these results confirm that \GR is not only better aligned with human scores globally, but also more reliable for within-group candidate selection.

\subsection{Validation of Key Design Choices}
\label{sec:ablations}

We keep the ablation study intentionally lightweight, since the main contribution of this work is the benchmark and the reward-model formulation rather than a complex architectural recipe. The goal of this section is to verify that the final configuration is supported by controlled development experiments.

\begin{table}[h]
\centering
\caption{Summary of key design-choice validations for \GR.}
\label{tab:ablation_summary}
\tablestyle{4pt}{1.1}
\resizebox{\columnwidth}{!}{
\begin{tabular}{l l l l}
\shline
Study & Compared settings & Selected choice & Observation \\
\shline
Loss function & REG / CLS / ORD & ORD & Best alignment with human labels \\
Temporal sampling & 1 / 2 / 4 / 8 FPS & 4 FPS & Best balance of motion and redundancy \\
Spatial resolution & 154K / 400K / 450K / 920K px & $\sim$400K px & Best trade-off between detail and efficiency \\
\shline
\end{tabular}
}
\vspace{-0.25cm}
\end{table}

\paragraph{Analysis.} Ordinal regression is consistently the strongest choice in development, which is well aligned with the ordered 1--4 label space. For video preprocessing, 4 FPS provides the best trade-off between temporal coverage and redundant frames. Spatially, around 400K pixels per frame is the most effective operating point: lower resolution removes subtle local editing cues, while higher resolution increases computation without yielding clear gains. These trends support the default configuration used in the final \GR models.

\section{Benchmarking Existing Video Editing Models}
\label{sec:benchmark_existing_models}

Beyond baseline comparison, \GB also enables a systematic evaluation of existing video editing models with our learned evaluator. We score 10 representative models using \GRl on the same 1--4 scale as \GD, including the commercial systems Kling o3 omni~\citep{kling2026video3omni}, Kling o1~\citep{kling2025o1}, Runway Gen-4.5~\citep{runway2025gen45}, Seedance 2.0~\citep{bytedance2026seedance2}, Grok Imagine~\citep{xai2026grokimagine}, Luma ray 3~\citep{luma2025ray3}, Wan 2.6~\citep{alibaba2025wan26}, and Luma ray 2~\citep{luma2025ray2}, as well as the open-source systems UniVideo~\citep{wei2025univideo} and VACE~\citep{jiang2025vace}. In this section, all reported metrics are computed from soft expected predictions for IF, RQ, and EE. We report \textit{Overall (Mean)} as the arithmetic mean of the three dimensions, and use \textit{Overall (GeoAgg)} as the primary ranking metric.

Following prior work on multiplicative multi-attribute aggregation, we define \textit{Overall (GeoAgg)} as a weighted geometric aggregate to reduce full compensability across dimensions and to penalize weak instruction following more strongly~\citep{keeney1993decisions,keeney1974multiplicative}. For each evaluated sample $i$ from model $m$, we first normalize the predicted scores to $[0,1]$:
\begin{equation}
i_{m,i} = \frac{IF_{m,i}-1}{3}, \qquad
r_{m,i} = \frac{RQ_{m,i}-1}{3}, \qquad
e_{m,i} = \frac{EE_{m,i}-1}{3}.
\end{equation}
We then compute the sample-level aggregate and average it over the evaluated set $\Omega_m$:
\begin{equation}
\label{eq:geoagg}
\text{Overall (GeoAgg)}_m =
\frac{1}{|\Omega_m|}\sum_{i\in\Omega_m}
\left[1 + 3\left(i_{m,i}^{\alpha} r_{m,i}^{\beta} e_{m,i}^{\gamma}\right)^{\frac{1}{\alpha+\beta+\gamma}}\right].
\end{equation}
In all experiments in this section, we set $(\alpha,\beta,\gamma)=(2,1,1)$, so IF receives twice the weight of RQ and EE. We compute GeoAgg before averaging because the geometric aggregate is nonlinear; applying it after averaging IF, RQ, and EE would overestimate systems with high variance or unbalanced per-sample behavior. Compared with an arithmetic mean, this multiplicative form is more sensitive to weak dimensions, which is desirable in video editing evaluation because strong rendering quality or locality preservation should not fully offset poor instruction following.

\paragraph{Adjusting incomplete model coverage.} Some commercial systems impose inference constraints, resulting in incomplete benchmark coverage for models such as Runway Gen-4.5 and Seedance 2.0. Rather than reporting a naive mean over each observed subset, which can be biased when coverage correlates with item difficulty, we treat incomplete coverage as a missing-data problem~\citep{rubin1976inference}. Our adjustment follows the standard inverse-propensity weighting principle~\citep{horvitz1952generalization,robins1994estimation,seaman2013review}. Let $R_{m,i}=1$ indicate that model $m$ has a valid evaluated output for benchmark item $i$, and let $\mathbf{x}_i$ denote item-level covariates such as task type, prompt length, and constraint count. We estimate the observation propensity $\hat{p}_{m,i}=\Pr(R_{m,i}=1\mid m,\mathbf{x}_i)$ and weight each observed score by a clipped inverse-propensity weight $w_{m,i}=1/\hat{p}_{m,i}$. For each dimension $d$, we then fit a weighted linear mixed-effects model
\begin{equation}
y_{m,i,d}=\mu_{m,d}+u_i+\epsilon_{m,i,d}, \qquad u_i \sim \mathcal{N}(0,\sigma_u^2),
\end{equation}
where $u_i$ captures item difficulty. The reported IF, RQ, and EE scores are coverage-adjusted model-level estimates $\hat{\mu}_{m,d}$ under the assumption that coverage is explainable by observed item covariates. \textit{Overall (Mean)} is computed from these adjusted dimension scores, while \textit{Overall (GeoAgg)} is computed from soft per-sample IF/RQ/EE predictions and then averaged as in \Cref{eq:geoagg}.

\begin{table}[h]
\centering
\caption{\GRl-based evaluation of representative video editing systems using soft expected predictions. IF, RQ, EE, and Overall (Mean) use coverage-adjusted estimates from inverse-propensity-weighted mixed-effects estimation; Overall (GeoAgg) is averaged over sample-level GeoAgg scores. Higher is better on all columns. $^\ast$ denotes adjusted results for models with incomplete benchmark coverage.}
\label{tab:editing_model_benchmark}
\tablestyle{4.5pt}{1.10}
\setlength{\tabcolsep}{15pt}
\resizebox{\columnwidth}{!}{
\begin{tabular}{l ccccc}
\shline
Model & Overall (GeoAgg) & Overall (Mean) & IF & RQ & EE \\
\shline
\multicolumn{6}{c}{\textit{Commercial}} \\
\hline
Kling o3 omni & \textbf{3.057} & \textbf{3.221} & \underline{3.033} & \textbf{3.588} & 3.043 \\
Kling o1 & \underline{2.985} & \underline{3.183} & \textbf{3.040} & \underline{3.534} & 2.976 \\
Runway Gen-4.5$^\ast$ & 2.912 & 3.020 & 2.817 & 3.319 & 2.923 \\
Seedance 2.0$^\ast$ & 2.766 & 3.107 & 2.811 & 3.421 & 3.088 \\
Grok Imagine & 2.723 & 3.109 & 2.606 & 3.346 & \textbf{3.376} \\
Luma ray 3 & 2.717 & 2.936 & 2.702 & 3.403 & 2.705 \\
Wan 2.6 & 2.146 & 2.592 & 2.012 & 3.317 & 2.446 \\
Luma ray 2 & 1.804 & 1.977 & 2.038 & 2.532 & 1.363 \\
\hline
\multicolumn{6}{c}{\textit{Open-source}} \\
\hline
UniVideo~\citep{wei2025univideo} & 2.516 & 2.883 & 2.294 & 3.266 & \underline{3.091} \\
VACE~\citep{jiang2025vace} & 1.775 & 2.126 & 2.027 & 3.172 & 1.180 \\
\shline
\end{tabular}
}
\vspace{-0.25cm}
\end{table}

\paragraph{Table analysis.} As shown in \Cref{tab:editing_model_benchmark}, Kling o3 omni ranks first under \textit{Overall (GeoAgg)}, followed by Kling o1. Both models combine strong IF and RQ with competitive EE, so their rankings remain high under the per-sample multiplicative aggregate. Runway Gen-4.5 ranks third by GeoAgg, reflecting balanced per-sample behavior despite a lower adjusted mean. Seedance 2.0 improves after the corrected result merge and ranks fourth by GeoAgg, with strong RQ and EE but still weaker IF than the top systems. Grok Imagine achieves the strongest EE score and a high arithmetic mean, but its lower IF reduces its \textit{Overall (GeoAgg)}.

Among the open-source systems, UniVideo is clearly stronger than VACE and remains competitive with several commercial systems, especially on EE. Luma ray 3 and Wan 2.6 achieve strong RQ but are limited by weaker IF or EE, while Luma ray 2 and VACE show the largest drops because of poor edit exclusivity. Overall, the results suggest that modern systems often produce visually plausible videos, but reliable instruction following and locality preservation still separate the strongest editing models from the rest.

\begin{figure}[h]
\centering
\includegraphics[width=\columnwidth]{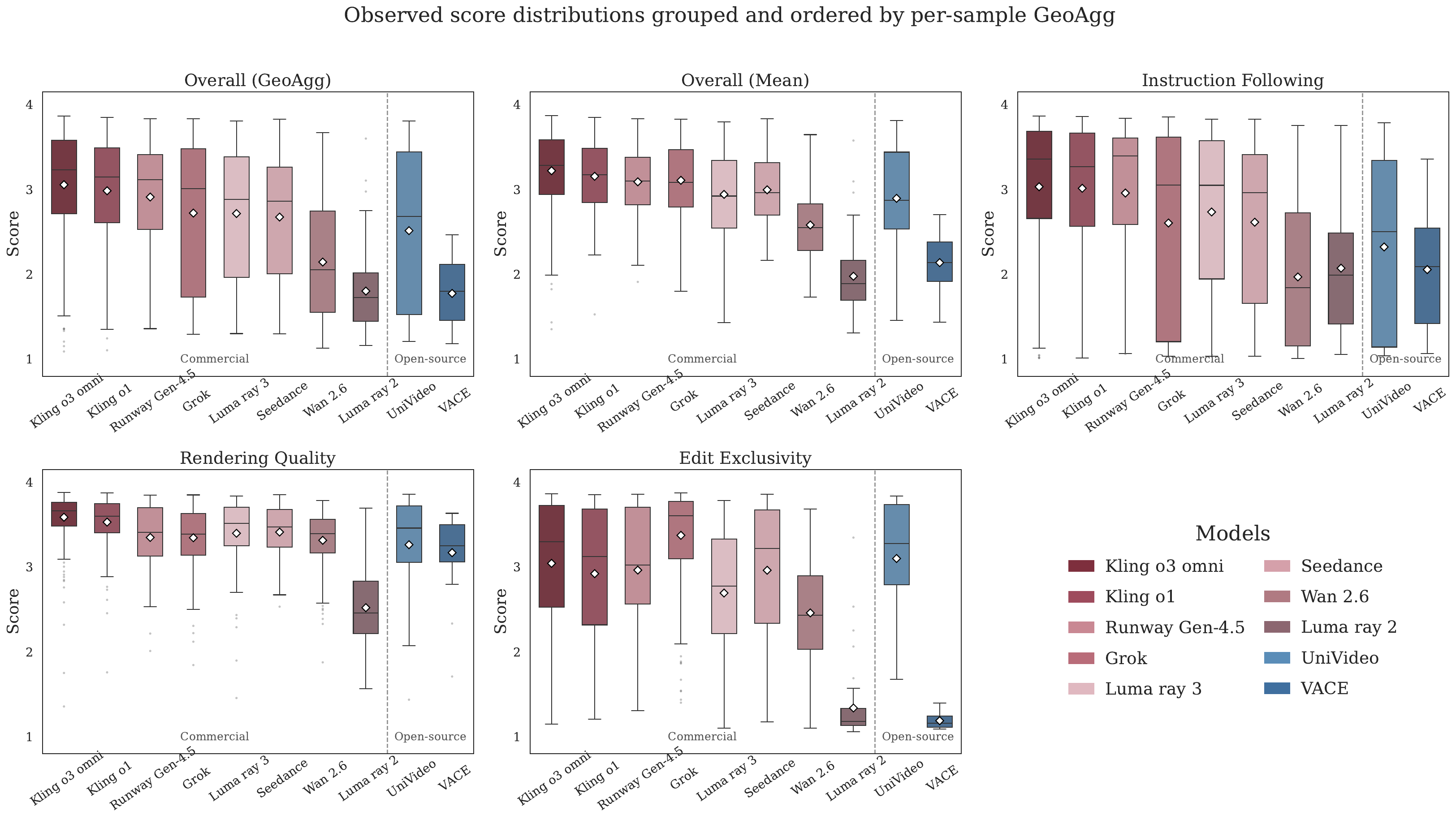}
\caption{Observed soft-score \GRl distributions for the benchmarked video editing systems across \textit{Overall (GeoAgg)}, \textit{Overall (Mean)}, IF, RQ, and EE. Models are grouped by availability and ordered by per-sample \textit{Overall (GeoAgg)}.}
\label{fig:editing_model_boxplot}
\vspace{-0.2cm}
\end{figure}

\paragraph{Figure analysis.} \Cref{fig:editing_model_boxplot} complements the adjusted table with the distribution of observed per-item scores. The top commercial systems have high medians but still show substantial prompt-level variance, indicating that no model is uniformly reliable across editing tasks. RQ is generally higher and more concentrated than IF, suggesting that visual plausibility is easier to achieve than instruction-faithful editing. EE provides the clearest separation: Grok Imagine, UniVideo, Kling o3 omni, and Seedance 2.0 maintain relatively strong locality, whereas VACE and Luma ray 2 concentrate near the bottom of the scale. The gap between \textit{Overall (Mean)} and \textit{Overall (GeoAgg)} is most visible for models with unbalanced dimensions, illustrating why a shortfall-sensitive aggregate is useful for benchmark ranking.

\begin{figure}[h]
\centering
\includegraphics[width=\textwidth]{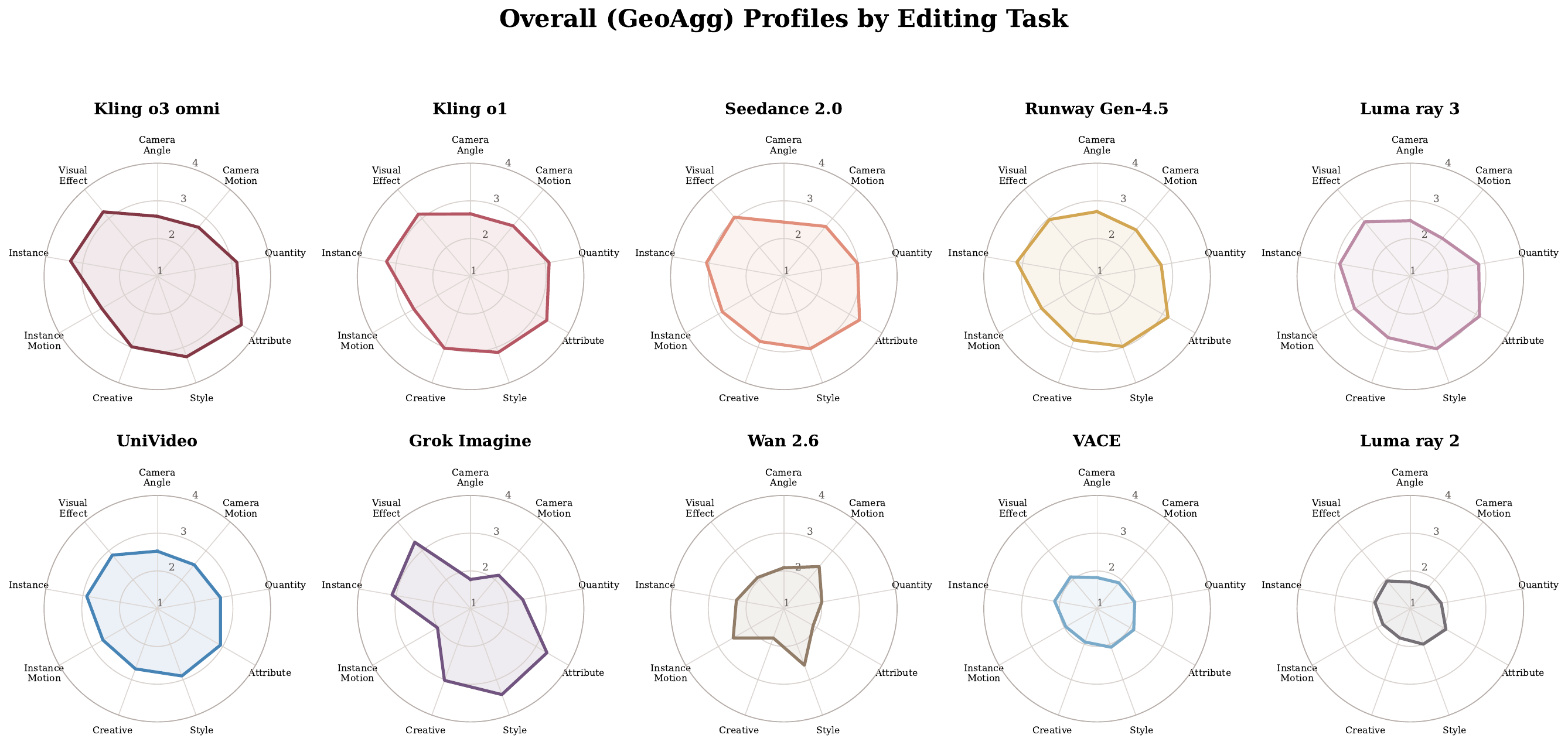}
\caption{Task-wise \textit{Overall (GeoAgg)} profiles of the benchmarked video editing systems. Each radar plot uses the same radial scale, allowing the profile shape and absolute score level of each model to be compared across editing tasks.}
\label{fig:editing_model_task_radar}
\vspace{-0.2cm}
\end{figure}

\paragraph{Task-wise analysis.} \Cref{fig:editing_model_task_radar} shows that the strongest systems are not uniformly strong across all editing types. Kling o3 omni and Kling o1 maintain broad coverage with clear advantages on quantity, attribute, instance, and visual-effect editing, while Runway Gen-4.5 and Seedance 2.0 are more balanced but slightly lower overall. Grok Imagine has a distinctive profile: it is strong on style, instance, and visual-effect editing, but weaker on camera-control tasks. The lower-scoring models show smaller and more compressed profiles, suggesting that their failures are not limited to a single task type.

\section{Conclusion}
\label{sec:conclusion}

We introduced \GD, a human-annotated dataset of 5,049 video editing examples with decoupled labels for Instruction Following, Rendering Quality, and Edit Exclusivity, together with \GR for automated evaluation and \GB for standardized model comparison. Across both standard IQA/VQA metrics and group-wise preference evaluation, \GR consistently outperforms generic VLM judges and prior reward-model baselines, showing the value of task-specific reward modeling for video editing. Using \GR as a scalable evaluator, we further benchmark representative commercial and open-source editing systems and analyze their behavior across editing tasks. This analysis shows that current systems often achieve plausible rendering quality without reliably satisfying instructions or preserving non-target content, reinforcing the need for multi-dimensional evaluation rather than a single holistic score. We hope these resources provide a practical foundation for benchmarking, model selection, and reward-driven optimization in video editing.

\newpage
{
\small
\bibliographystyle{IEEEtran}
\bibliography{main}
}

\clearpage
\newpage
\beginappendix
\section{Additional Training Details for \GR}
\label{sec:appendix_training_details}

We provide the implementation details omitted from the main paper. \GR is trained on the 4,200-example training split of \GD and evaluated on the 849-example test split, with the split stratified across editing categories and pipelines.

\paragraph{Video Processing.} For each example, we uniformly sample both the original and edited videos at 4 FPS and cap the frame resolution at 399,360 pixels, approximately 632 $\times$ 632, while preserving aspect ratio through Qwen3-VL's dynamic-resolution mechanism. The two videos are sampled with aligned temporal indices to support direct comparison. The maximum sequence length is 32,768 tokens.

\paragraph{Optimization.} We use a two-stage training schedule. In the first stage, lasting 1 epoch, we freeze all pretrained parameters and train only the newly introduced reward tokens and reward head. In the second stage, lasting 49 epochs, we unfreeze and fine-tune the language backbone and visual-language merger together with the reward head and reward tokens, while keeping the vision tower frozen. We use AdamW with learning rates of $1 \times 10^{-5}$ for the language-side parameters and $5 \times 10^{-5}$ for the reward tokens, cosine decay, and a 15\% warmup ratio. Training is performed in bf16 on 8 GPUs with an effective batch size of 8. The three reward dimensions are optimized jointly with equal loss weights.

\section{Additional Experimental Details}
\label{sec:appendix_experiment_details}

\paragraph{Model Variants.} We evaluate two \GR variants, \GRs and \GRl, which instantiate the same reward-model design on Qwen3-VL backbones at 4B and 32B scales.

\paragraph{Evaluation Setup.} Both variants are evaluated on the same 849-example test split. For VLM-as-judge baselines, we use a shared prompt that presents the original video, editing instruction, and edited video, and asks the model to score IF, RQ, and EE according to the same 1--4 rubric used in human annotation. In the main paper, the human overall score is defined as the arithmetic mean of the three human dimension scores. For \GR and VLM-as-judge baselines, the overall prediction is defined as the mean of the three predicted dimension scores.

\paragraph{External Reward Models.} EditReward and VE-Bench are evaluated through their native outputs. For EditReward, the overall prediction is defined as the mean of its two native heads; for VE-Bench, the overall prediction is its native scalar output. SRCC and KRCC are computed directly from raw predictions, while PLCC and RMSE use the same logistic calibration protocol as in the main paper when applicable. Since EditReward lacks a dedicated EE head and VE-Bench predicts only a single overall score, unavailable entries are marked with \ldash.

\section{Editing Pipeline Details}
\label{sec:appendix_pipelines}

We describe the detailed procedures for each category of editing systems used in \GD.

\subsection{Commercial Models}

For generic instruction-guided video editing, we directly submit (source video, instruction) pairs to four commercial APIs: Grok Imagine~\citep{xai_grok_imagine_2026}, Kling Omni~\citep{klingai_omninew_2025}, Wan 2.6~\citep{cheng2025wan}, and Luma Ray2~\citep{lumaray2_2025}. These systems accept free-form text instructions and produce edited videos end-to-end.

\subsection{Open-Source Specialized Models}

\paragraph{Instance Removal.} We first apply SAM 2~\citep{ravi2024sam2} to segment the target instance across all video frames, with manual verification and correction of segmentation masks. The corrected masks are then fed to ROSE~\citep{miao2025rose}, using the removal model retrained on the ROSE dataset with the PISCO~\citep{gao2026pisco} framework to support 720p resolution and 121-frame sequences.

\paragraph{Instance Insertion.} We use NanoBanana-Pro to perform the desired edit on a single reference frame, then extract the newly inserted object via SAM 2 segmentation. The extracted single-frame instance serves as a spatial control signal for PISCO, which propagates the insertion consistently across all video frames while preserving the background.

\paragraph{Instance Repositioning and Resizing.} The target instance is extracted using SAM 2 with manual correction. A VLM (Gemini-2.0-Flash~\citep{deepmind_gemini_flash_2025}) interprets the editing instruction and provides guidance for the required spatial transformation (translation, scaling, rotation) of the segmented instance. Simultaneously, the PISCO-finetuned removal model inpaints the vacated region. Finally, PISCO-14B performs instance insertion using the transformed instance as a spatial condition, composited onto the inpainted background video.

\paragraph{Human Motion Editing.} We extract human pose keypoints using ViTPose~\citep{xu2022vitpose} and modify them according to Gemini-2.0-Flash's interpretation of the editing instruction, with human-in-the-loop verification of pose correctness. The modified pose sequence, together with additional control signals (Canny edges, depth maps from Depth Anything V3~\citep{lin2025depth}), condition Wan-Animate to generate a new video from the original first frame.

\paragraph{Camera Motion and Angle Editing.} Gemini-2.0-Flash maps the natural language instruction to predefined camera trajectory parameters (pan, tilt, zoom, dolly, arc, etc.). ReCamMaster~\citep{he2024recammaster} and LightX~\citep{liu2025light} then execute the specified camera transformation on the source video.

\paragraph{Style, Creative, Visual Effect, and Attribute Editing.} For these categories, we apply NanoBanana-Pro to edit the first frame according to the instruction, then use VACE~\citep{jiang2025vace} in first-frame-conditioned mode to propagate the edit temporally across all frames. Additionally, for a subset of samples, we employ UniVideo~\citep{wei2025univideo} for direct end-to-end text-conditioned video editing, similar to the commercial-model usage pattern.

\section{Complete Annotation Guide}
\label{sec:appendix_guide}

This appendix provides the complete annotation guide used to train annotators for \GD. The guide was provided in both English and Chinese; we present the English version here.

\subsection{General Instructions}

Annotators are presented with an original video, an editing instruction, and one or more edited videos produced by different models. For each edited video, annotators independently score three dimensions on a 4-point scale (1--4). The three dimensions must be scored independently: the score on one dimension must not influence the score on another. When a result matches multiple descriptions, annotators should assign the lowest applicable score. This rule is especially important when a video has both good aspects and one clear failure that crosses the boundary to a lower level.

\subsection{Dimension 1: Instruction Following (IF)}

This dimension evaluates whether the edited content accurately reflects the semantic requirements of the instruction.

\begin{itemize}[leftmargin=*, itemsep=2pt]
    \item \textbf{Score 4 --- Complete and Correct Execution.} All requested edits are clearly completed, and no required element is missing or incorrect. The target object, attribute, action, style, or camera change matches the instruction without visible contradiction.

    \item \textbf{Score 3 --- Mostly Correct Execution.} The core edit is completed, but one minor detail is wrong or missing. Typical cases include correct target and edit type but slight mismatch in fine-grained attribute, appearance, intensity, or local extent. The result should still be recognizably aligned with the instruction overall.

    \item \textbf{Score 2 --- Partial Execution with Major Deviation.} The video shows some relationship to the instruction, but the main requirement is only partially satisfied or is satisfied with a major semantic error. Typical cases include editing the correct region but producing the wrong object or attribute, executing only one part of a multi-step instruction, or mixing the requested edit with an obviously incorrect alternative.

    \item \textbf{Score 1 --- Failure or Contradiction.} The instruction is not executed, the edit is largely unrelated, or the result directly contradicts the instruction. Examples include no visible edit, editing the wrong target, or changing the scene in the opposite direction from the requested operation.
\end{itemize}

\subsection{Dimension 2: Rendering Quality (RQ)}

This dimension evaluates the visual quality of the edited video, including naturalness, clarity, physical correctness of object movements, temporal consistency between frames, and the absence of artifacts.

\begin{itemize}[leftmargin=*, itemsep=2pt]
    \item \textbf{Score 4 --- High Visual Fidelity.} The video is clear, temporally stable, and visually natural throughout. Artifacts are absent or only barely perceptible, object structure remains intact, and motion follows plausible physical behavior.

    \item \textbf{Score 3 --- Minor but Noticeable Degradation.} The video remains fully watchable, but there are visible quality issues such as slight blur, local flicker, mild temporal inconsistency, or small artifact regions. These issues are limited and do not damage the overall scene structure or object identity.

    \item \textbf{Score 2 --- Clear Quality Failure.} Artifacts are obvious and recurrent, such as repeated flicker, deformation, ghosting, severe blur, unstable boundaries, or unnatural motion. The content is still recognizable, but the defects substantially reduce visual quality and viewing coherence.

    \item \textbf{Score 1 --- Severe Visual Breakdown.} The result is visually unusable or close to unusable. Major regions are corrupted, object identity collapses, temporal coherence is lost, or motion becomes physically implausible to the point that the video no longer supports reliable evaluation of the intended edit.
\end{itemize}

\subsection{Dimension 3: Edit Exclusivity (EE)}

This dimension evaluates whether the model executed only the specified operation without unnecessary changes to unrelated areas. A non-target change is defined as any clearly visible modification to an object, region, or background element that is not required by the instruction. When counting non-target changes, multiple altered instances in different semantic regions should be counted separately.

\begin{itemize}[leftmargin=*, itemsep=2pt]
    \item \textbf{Score 4 --- Strict Preservation.} No clearly visible non-target change is introduced. All regions outside the intended edit remain visually unchanged, except for imperceptible pixel-level differences or negligible rendering noise.

    \item \textbf{Score 3 --- One Clear Non-Target Change.} The intended target is edited, but exactly one additional non-target object or semantic region is also clearly altered. The overall scene layout is still preserved, and the error remains localized.

    \item \textbf{Score 2 --- Two to Three Clear Non-Target Changes.} Two or three non-target objects or semantic regions are clearly altered, or one large unintended background change affects a substantial part of the scene. The result still resembles the original video, but over-editing is obvious.

    \item \textbf{Score 1 --- Global or Widespread Over-Editing.} More than three non-target objects or semantic regions are clearly altered, or the scene is globally rewritten. The result looks like a substantially different video rather than a localized edit.
\end{itemize}

\subsection{Dimension Decoupling Principle}

The three dimensions must be scored independently. Consider the following example:

\paragraph{Instruction:} ``Turn the apple into a banana.''

\paragraph{Result:} The model completely fails and the apple remains unchanged.

\begin{itemize}[leftmargin=*, itemsep=1pt]
    \item IF = 1 (complete failure to follow the instruction)
    \item RQ = score independently (if the video quality is excellent, this can still be 4)
    \item EE = score independently (if no unintended changes occurred, this can still be 4)
\end{itemize}

This principle ensures that each dimension captures a distinct aspect of editing quality.

\subsection{Annotation Examples}
\label{sec:appendix_examples}

We provide all example cases from the annotation guide. Each figure shows the first frame of the original video together with one or more edited results and their IF/RQ/EE scores. The scores are assigned from the full video rather than from the displayed frame alone. These examples cover attribute editing, creative editing, instance editing, visual effects, and style transfer, and illustrate how the same instruction can lead to different score patterns across dimensions.

\begin{figure*}[h]
\centering
\includegraphics[width=\textwidth]{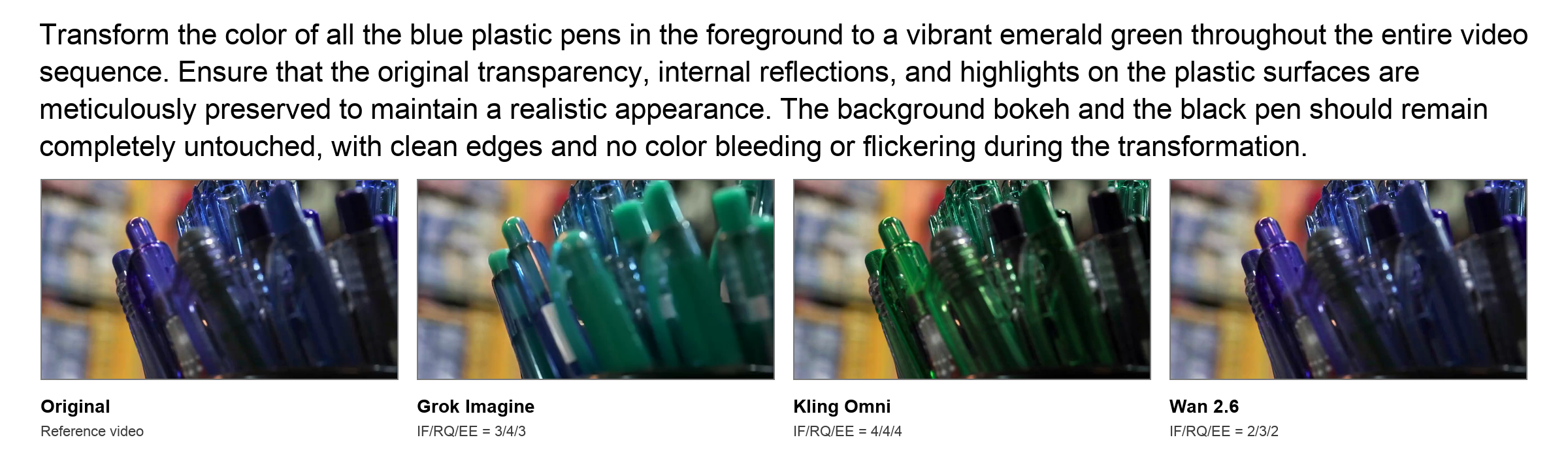}
\caption{Annotation example 1. The instruction asks to turn only the blue foreground pens into emerald green while preserving transparency, reflections, highlights, the black pen, and the blurred background. Kling Omni receives IF/RQ/EE $=4/4/4$ because it executes the requested color change cleanly, keeps the plastic appearance realistic, and leaves non-target content untouched. Grok Imagine receives $3/4/3$ because the target edit is mostly correct and visually clean, but the green conversion is less precise and some non-target regions are also affected, reducing both IF and EE. Wan 2.6 receives $2/3/2$ because the requested color transformation is incomplete, the result looks less stable and less realistic, and unintended color changes spill into regions that should have remained unchanged.}
\label{fig:annotation_example_1}
\end{figure*}

\begin{figure*}[h]
\centering
\includegraphics[width=\textwidth]{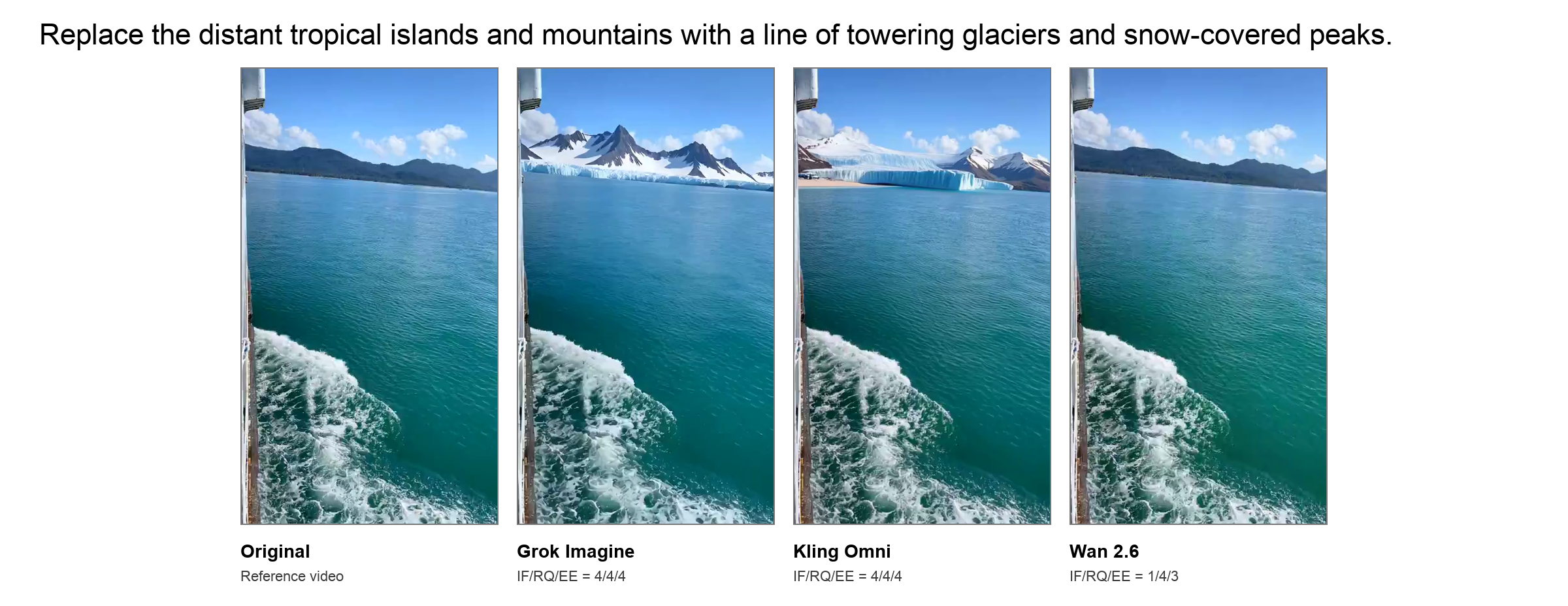}
\caption{Annotation example 2. The instruction asks to replace the distant tropical islands and mountains with glaciers and snow-covered peaks while leaving the rest of the scene intact. Grok Imagine and Kling Omni both receive IF/RQ/EE $=4/4/4$ because they fully carry out the requested background replacement, keep the water and boat-view foreground natural, and introduce no obvious non-target distortions. Wan 2.6 receives $1/4/3$ because it essentially fails to perform the requested semantic edit: the original tropical background remains, so IF is 1. Its RQ is still 4 because the video itself remains visually clean and artifact-free, which illustrates the intended decoupling between instruction following and rendering quality. EE is 3 rather than 4 because, although the edit does not heavily corrupt the scene, the output also does not faithfully realize the target modification.}
\label{fig:annotation_example_2}
\end{figure*}

\begin{figure*}[h]
\centering
\includegraphics[width=\textwidth]{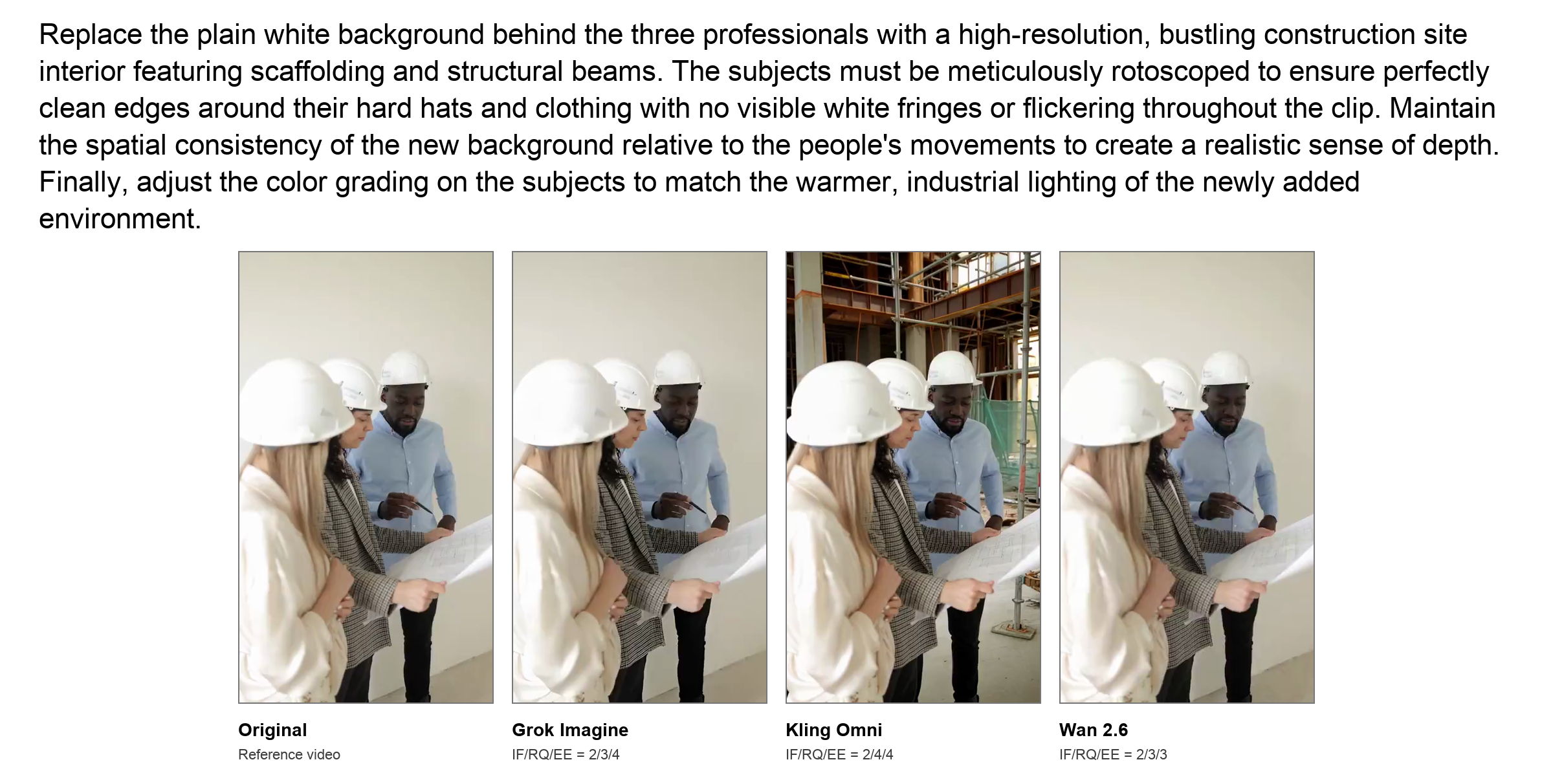}
\caption{Annotation example 3. The instruction asks for a complete replacement of the white background with a bustling construction-site interior, together with clean subject boundaries, stable depth, and relit foreground subjects. All three edited results receive IF $=2$ because they only partially satisfy the instruction: the construction-site replacement is introduced, but the compositing and relighting are not fully convincing, so the overall request is only partly achieved. Kling Omni receives the highest rendering score, RQ $=4$, because its compositing is the cleanest and most visually coherent over time. Grok Imagine and Wan 2.6 receive RQ $=3$ because the inserted environment looks less seamlessly integrated and shows weaker consistency. Grok Imagine and Kling Omni both receive EE $=4$ because the three people remain largely intact, whereas Wan 2.6 receives EE $=3$ because the foreground subjects are altered more noticeably during the edit.}
\label{fig:annotation_example_3}
\end{figure*}

\begin{figure*}[h]
\centering
\includegraphics[width=\textwidth]{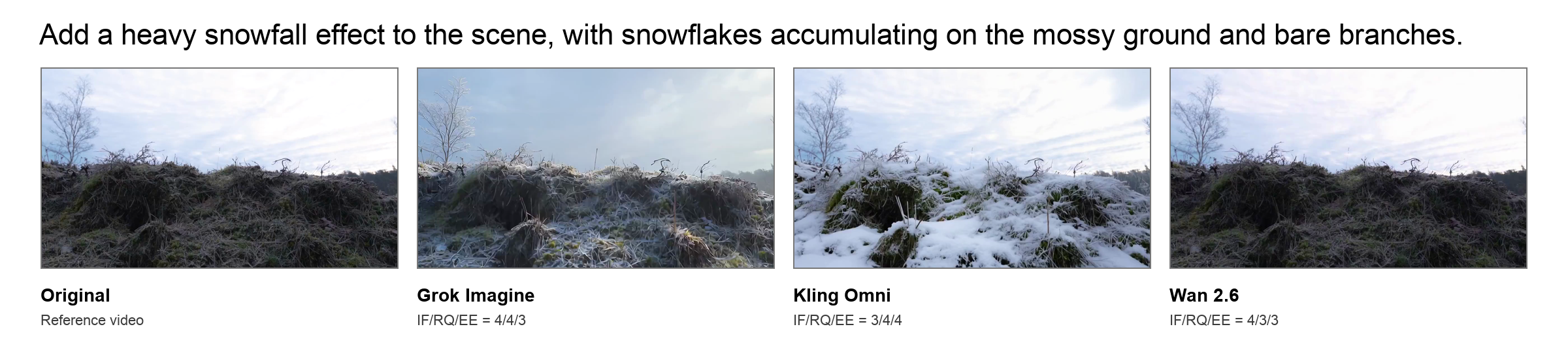}
\caption{Annotation example 4. The instruction asks for a heavy snowfall effect with visible accumulation on the mossy ground and bare branches. Grok Imagine receives IF/RQ/EE $=4/4/3$ because it clearly adds snow and accumulation with strong visual quality, but it also alters parts of the scene beyond the requested effect. Kling Omni receives $3/4/4$ because the result is visually strong and preserves the scene structure well, but the snowfall effect is weaker than requested, so the instruction is not fully satisfied. Wan 2.6 receives $4/3/3$ because it does introduce the requested snow effect, but the rendering is less realistic and less stable, and some non-target structure is also modified.}
\label{fig:annotation_example_4}
\end{figure*}

\clearpage

\begin{figure*}[h]
\centering
\includegraphics[width=\textwidth]{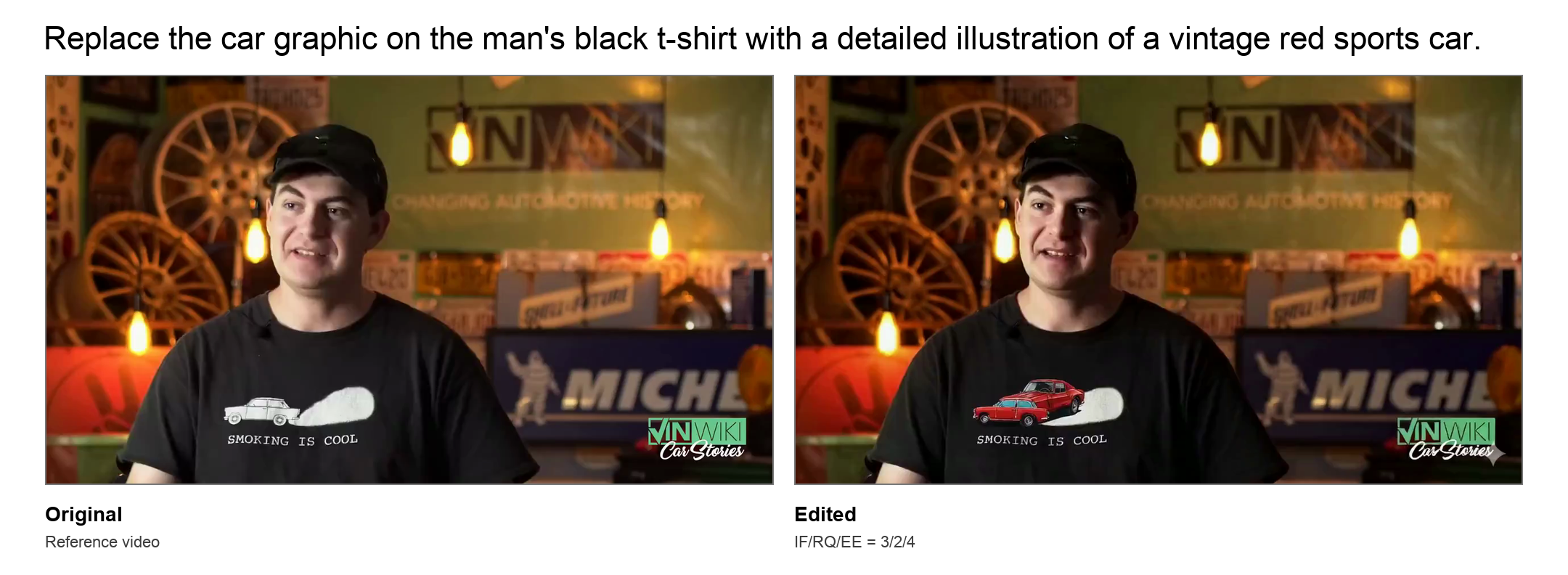}
\caption{Annotation example 5. The instruction asks to replace the shirt graphic with a detailed vintage red sports car while keeping the rest of the person and scene unchanged. The edited result receives IF/RQ/EE $=3/2/4$. IF is 3 because the shirt graphic is changed to a red car, so the main semantic request is met, but the inserted graphic is not fully convincing as a detailed vintage illustration across the clip. RQ is 2 because the edited graphic shows noticeable temporal instability and tracking inconsistency over time, even though the displayed frame looks acceptable. EE is 4 because the edit remains well localized to the shirt and does not introduce obvious unintended changes elsewhere in the video.}
\label{fig:annotation_example_5}
\end{figure*}

\begin{figure*}[h]
\centering
\includegraphics[width=\textwidth]{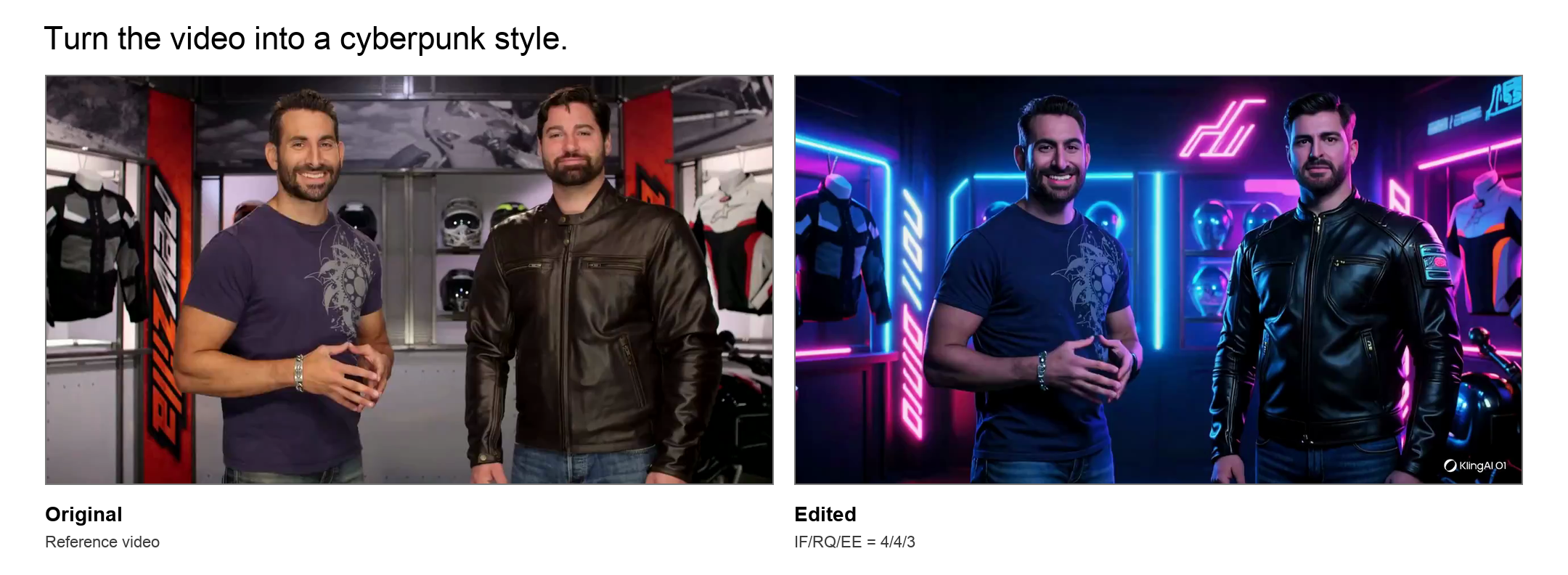}
\caption{Annotation example 6. The instruction asks to convert the video into a cyberpunk style. The edited result receives IF/RQ/EE $=4/4/3$. IF is 4 because the neon lighting, color palette, wardrobe styling, and overall atmosphere clearly match the requested cyberpunk aesthetic. RQ is 4 because the stylization is visually coherent and clean. EE is reduced to 3 because the transformation also modifies unrelated details, including text, facial appearance, and other local elements beyond the minimal style change needed to satisfy the instruction.}
\label{fig:annotation_example_6}
\end{figure*}

\begin{figure*}[h]
\centering
\includegraphics[width=\textwidth]{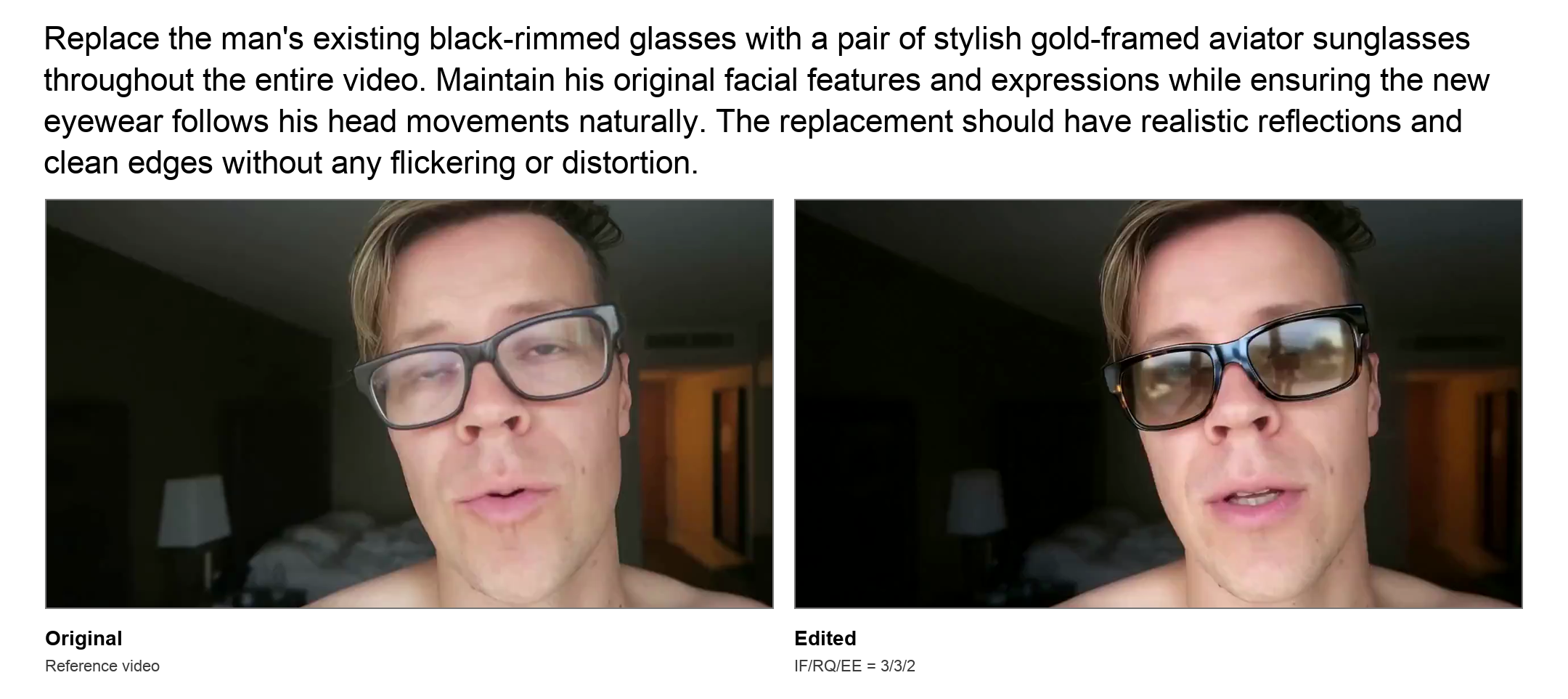}
\caption{Annotation example 7. The instruction asks to replace the original black-rimmed glasses with gold-framed aviator sunglasses while preserving the man's facial features, expressions, head motion, reflections, and clean edges. The edited result receives IF/RQ/EE $=3/3/2$. IF is 3 because the target object is indeed changed into sunglasses, but the replacement does not fully satisfy the requested appearance and realism. RQ is 3 because the local edit is usable but not fully clean, with only moderate realism in the eyewear integration. EE is 2 because the edit also changes other facial details, overall lighting, and background appearance, producing multiple unintended modifications outside the requested eyewear replacement.}
\label{fig:annotation_example_7}
\end{figure*}

\section{Inter-Annotator Agreement}
\label{sec:appendix_iaa}

\subsection{Cross-Check Procedure}

We randomly sample 550 examples from the annotated dataset and assign them to an independent group of new annotators for re-annotation. The second group follows the same annotation protocol and training procedure but has no access to the original annotations. This produces a double-annotation subset for a focused consistency check.

\subsection{Agreement by Dimension}

\Cref{tab:iaa_detailed} reports the agreement statistics used in the main paper.

\begin{table}[h]
\centering
\caption{Inter-annotator agreement on the 550-sample double-annotation subset.}
\label{tab:iaa_detailed}
\vspace{0.2cm}
\tablestyle{6pt}{1.15}
\begin{tabular}{l c c c}
\shline
Metric & IF & RQ & EE \\
\shline
Exact Agreement (\%) & 75.2 & 87.2 & 72.2 \\
Within-1 Agreement (\%) & 93.5 & 97.2 & 91.7 \\
\shline
\end{tabular}
\end{table}

RQ achieves the strongest agreement, with 87.2\% exact agreement and 97.2\% within-1 agreement, indicating that rendering quality is relatively stable across annotators. IF also shows strong consistency, with 75.2\% exact agreement and 93.5\% within-1 agreement. EE remains the most challenging dimension, but still reaches 72.2\% exact agreement and 91.7\% within-1 agreement, which suggests that judgments about non-target changes are noisier yet still broadly consistent. Overall, these results support the reliability of the three-dimensional annotation protocol while also reflecting the inherently subjective nature of fine-grained video-editing assessment.

\section{Extended Dataset Analysis}
\label{sec:appendix_analysis}

We present additional analyses of \GD that complement the main text.

\subsection{Task Type Difficulty Ranking}

\begin{figure}[h!]
\centering
\includegraphics[width=0.9\columnwidth]{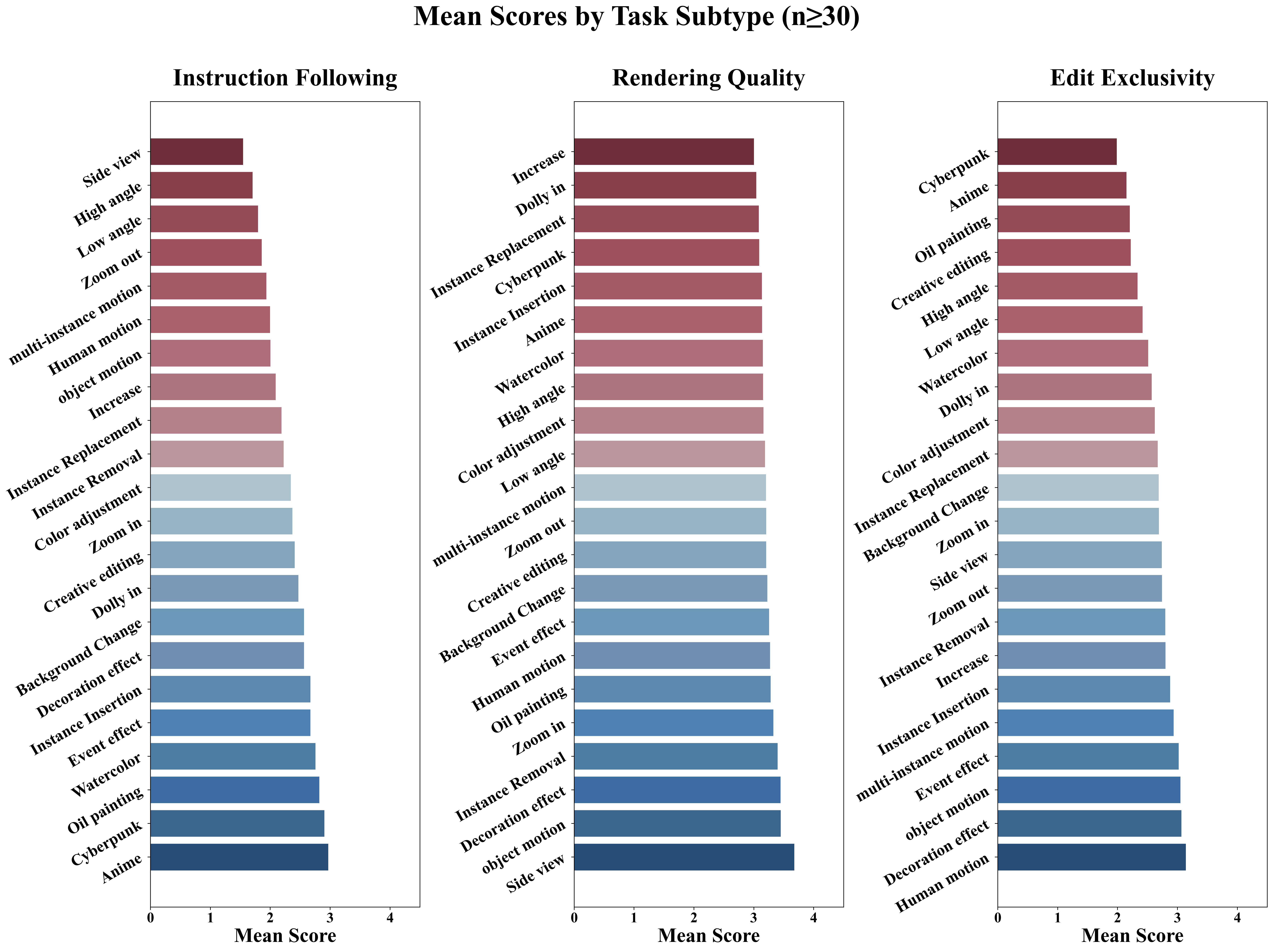}
\caption{Score breakdown across all 32 subcategories grouped by 9 main categories. Fine-grained subcategory variation reveals which specific editing operations are most challenging.}
\label{fig:task_subtype}
\end{figure}

\Cref{tab:task_difficulty} ranks the 9 task types by IF difficulty, and \Cref{fig:task_subtype} provides a finer-grained view across all 32 subcategories.

\begin{table}[h]
\centering
\caption{Task type ranking by editing difficulty. Tasks are sorted by IF score in ascending order, so lower values indicate harder semantic execution.}
\label{tab:task_difficulty}
\vspace{0.2cm}
\tablestyle{3pt}{1.15}
\begin{tabular}{l c c c c}
\shline
Task Type & IF & RQ & EE & $N$ \\
\shline
Camera Angle & 1.76 & 3.20 & 2.41 & 796 \\
Instance Motion & 2.00 & 3.39 & 3.06 & 450 \\
Quantity & 2.09 & 3.00 & 2.81 & 634 \\
Camera Motion & 2.31 & 3.28 & 2.69 & 383 \\
Attribute & 2.35 & 3.16 & 2.63 & 598 \\
Creative & 2.41 & 3.21 & 2.22 & 542 \\
Instance & 2.51 & 3.14 & 2.82 & 641 \\
Visual Effect & 2.58 & 3.32 & 2.89 & 520 \\
Style & 2.87 & 3.14 & 2.23 & 485 \\
\shline
\end{tabular}
\end{table}

Camera Angle editing is the hardest task for IF, as it requires geometric and 3D scene reasoning that current systems still handle poorly. Style Editing is the easiest for IF but has relatively low EE, reflecting the inherent tension between global style transformation and strict locality preservation. Notably, RQ varies much less across task types than IF or EE, which again suggests that current models find visual plausibility easier than precise semantic execution.

\subsection{Video Difficulty and Training Signal Quality}

\begin{figure}[h]
\centering
\includegraphics[width=0.9\columnwidth]{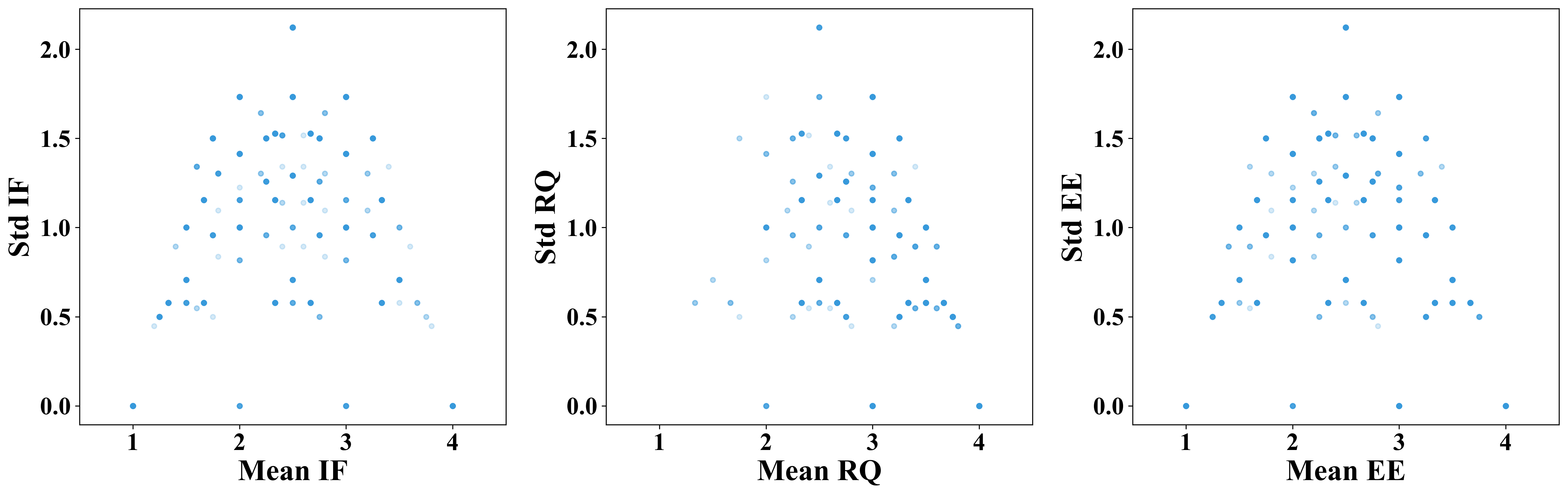}
\caption{Per-video difficulty (mean score) versus cross-pipeline score variance. High-variance videos provide especially informative supervision for reward modeling.}
\label{fig:video_difficulty}
\end{figure}

\Cref{fig:video_difficulty} shows the distribution of per-video difficulty, measured as the mean score across pipelines, against cross-pipeline score variance. Videos with high score variance are especially valuable for reward-model learning because they provide strong preference signals: different pipelines succeed or fail on the same input, enabling the reward model to learn discriminative features rather than a dataset-wide average.

\section{Per-Category Detailed Results}
\label{sec:appendix_per_category}

We provide finer-grained results for the six representative editing systems benchmarked in \Cref{sec:benchmark_existing_models}. Each heatmap-style table reports the mean \GRl score for one dimension at the level of the 9 main editing categories, using the same model ordering as the main-text benchmark. To keep the focus on comparative behavior rather than coverage statistics, we intentionally omit sample-count details here.

Across the 9 main categories, IF shows the largest variation and remains the main source of separation between models. Grok Imagine and Kling Omni are strongest on many attribute, style, and instance-editing tasks, while camera-angle and camera-motion edits remain difficult for nearly all systems. RQ is comparatively stable across categories, indicating that visually plausible outputs are often easier to produce than semantically correct ones. EE reveals the sharpest locality gap: Grok Imagine and UniVideo remain relatively strong on localized edits, whereas VACE and Luma Ray2 degrade more visibly when preserving non-target regions becomes difficult.

\begin{figure*}[h]
\centering
\includegraphics[width=0.92\textwidth]{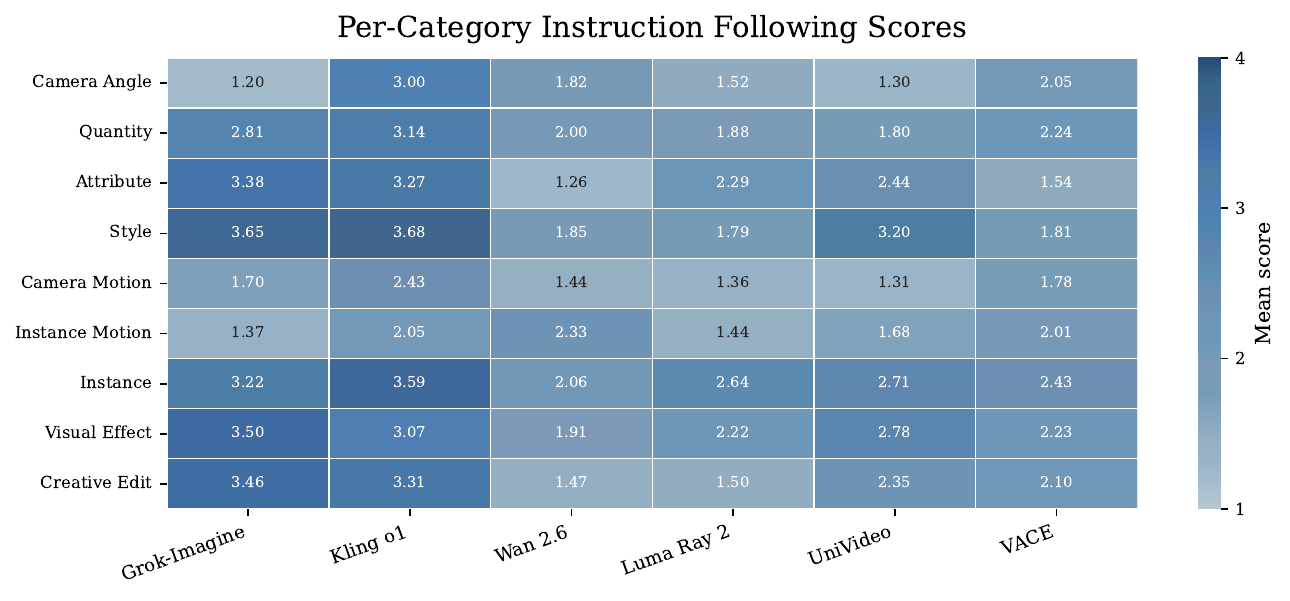}
\caption{Heatmap table of per-category Instruction Following scores for the six benchmarked video editing systems. Darker colors indicate higher \GRl scores.}
\label{fig:per_subcategory_if}
\end{figure*}

\begin{figure*}[h]
\centering
\includegraphics[width=0.92\textwidth]{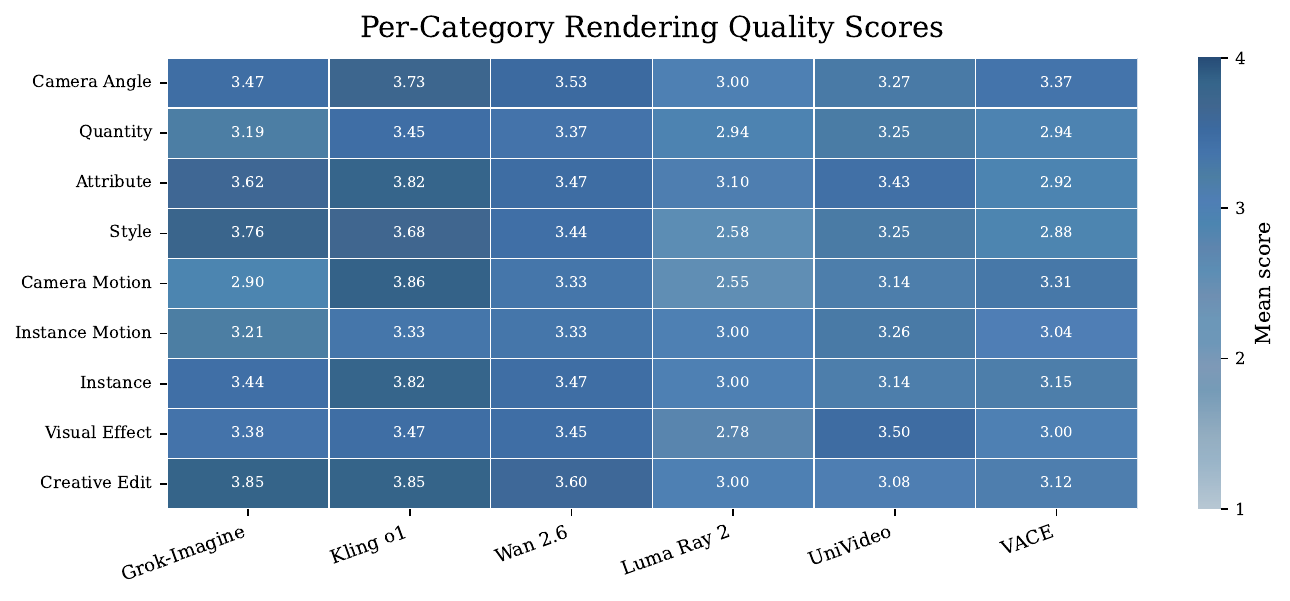}
\caption{Heatmap table of per-category Rendering Quality scores for the six benchmarked video editing systems.}
\label{fig:per_subcategory_rq}
\end{figure*}

\begin{figure*}[h]
\centering
\includegraphics[width=0.92\textwidth]{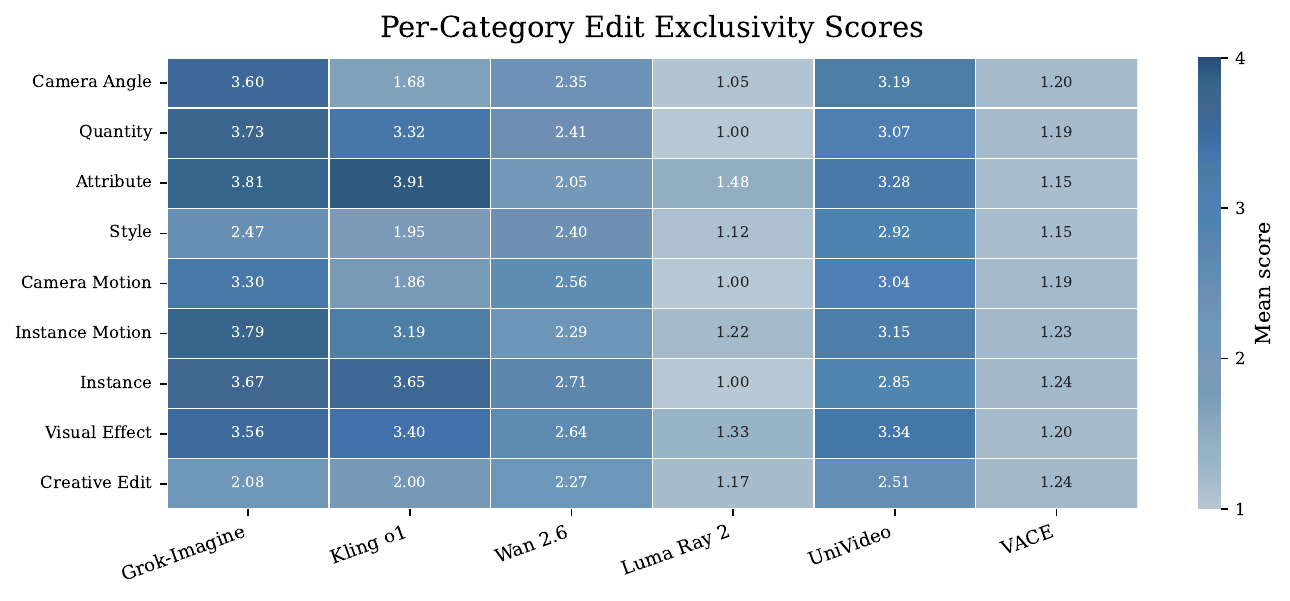}
\caption{Heatmap table of per-category Edit Exclusivity scores for the six benchmarked video editing systems.}
\label{fig:per_subcategory_ee}
\end{figure*}

\section{Additional Evaluation Details}
\label{sec:appendix_eval_details}

\subsection{Definitions of Standard IQA/VQA Metrics}
\label{sec:appendix_classical_metrics}

In \Cref{sec:classical_metrics}, we report four standard IQA/VQA-style metrics. We summarize their definitions here for completeness.

\paragraph{Spearman Rank-Order Correlation Coefficient (SRCC).} SRCC measures monotonic agreement between predicted scores and human labels:
\begin{equation}
\mathrm{SRCC} = 1 - \frac{6 \sum_{i=1}^{n} d_i^2}{n(n^2 - 1)},
\end{equation}
where $d_i$ is the rank difference between the $i$-th prediction and the corresponding human score.

\paragraph{Kendall Rank-Order Correlation Coefficient (KRCC).} We use Kendall's $\tau$-b to account for ties in the discrete 1--4 human scores:
\begin{equation}
\tau = \frac{N_c - N_d}{\sqrt{\left(\frac{n(n-1)}{2} - T_{\mathrm{pred}}\right)\left(\frac{n(n-1)}{2} - T_{\mathrm{human}}\right)}},
\end{equation}
where $N_c$ and $N_d$ denote the numbers of concordant and discordant pairs, and $T_{\mathrm{pred}}$ and $T_{\mathrm{human}}$ denote the numbers of tied pairs in the predicted and human rankings.

\paragraph{Pearson Linear Correlation Coefficient (PLCC).} PLCC measures linear agreement after score calibration:
\begin{equation}
\mathrm{PLCC} = \frac{\sum_{i=1}^{n} (x_i - \bar{x})(y_i - \bar{y})}{\sqrt{\sum_{i=1}^{n} (x_i - \bar{x})^2 \sum_{i=1}^{n} (y_i - \bar{y})^2}},
\end{equation}
where $x_i$ is the calibrated model prediction and $y_i$ is the human score.

\paragraph{Root Mean Squared Error (RMSE).} RMSE measures the calibrated absolute deviation:
\begin{equation}
\mathrm{RMSE} = \sqrt{\frac{1}{n}\sum_{i=1}^{n}(x_i - y_i)^2}.
\end{equation}

\paragraph{Logistic calibration.} Following common IQA/VQA protocol, we apply a four-parameter logistic mapping before computing PLCC and RMSE:
\begin{equation}
q(x) = \beta_1 \left(\frac{1}{2} - \frac{1}{1 + e^{\beta_2 (x - \beta_3)}}\right) + \beta_4,
\end{equation}
where $x$ is the raw model score and $q(x)$ is the calibrated score. The parameters $\beta_1,\dots,\beta_4$ are fitted by non-linear least squares on the evaluation set.

\begin{figure}[h]
\centering
\includegraphics[width=\columnwidth]{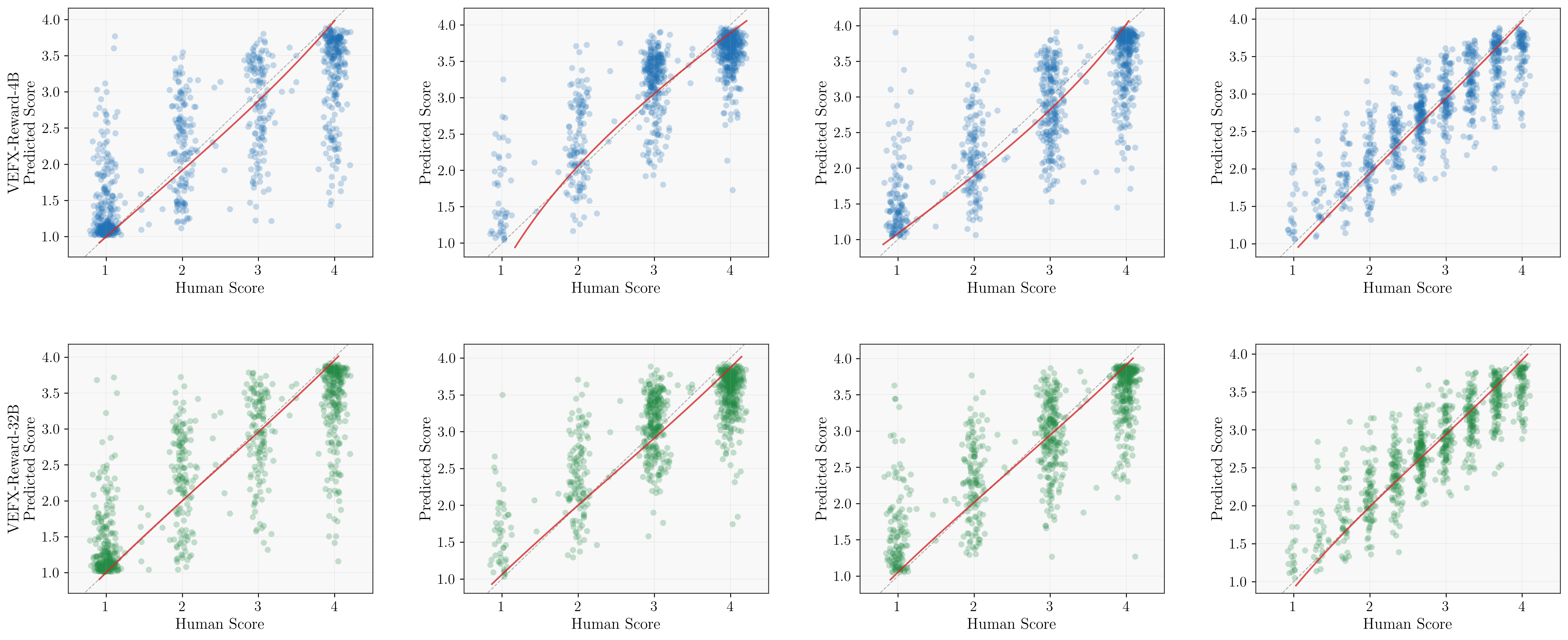}
\caption{Side-by-side comparison between \GRs and \GRl across IF, RQ, EE, and Overall.}
\label{fig:scatter_4b_32b_appendix}
\end{figure}

\paragraph{Supplementary scatter plot.} \Cref{fig:scatter_4b_32b_appendix} directly compares \GRs and \GRl across IF, RQ, EE, and Overall. Here Overall is defined as the mean of IF, RQ, and EE for both predictions and human scores. The figure shows that scaling from 4B to 32B mainly improves IF, EE, and the overall score, where the 32B predictions form visibly tighter trends around the human annotations. By contrast, RQ remains relatively similar across scales, consistent with the main-text observation that rendering quality is already easier to model than semantic faithfulness and edit locality.

\subsection{Additional Benchmark Visualizations}
\label{sec:appendix_model_benchmark}

\begin{figure}[!h]
\centering
\includegraphics[width=\columnwidth]{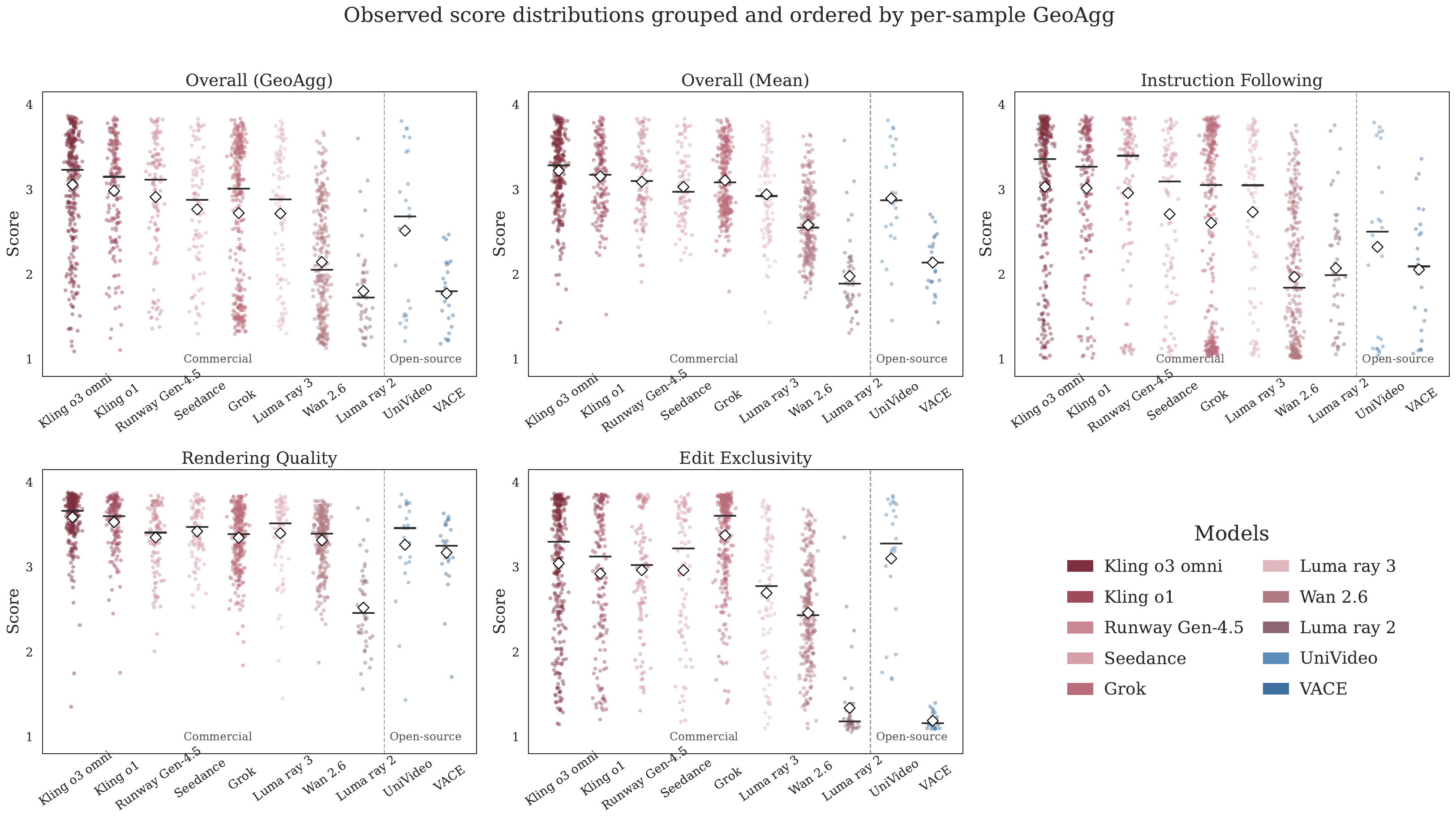}
\caption{Strip plot of the six-model evaluation, showing individual \GRl scores for IF, RQ, and EE.}
\label{fig:benchmark_strip_appendix}
\end{figure}

\begin{figure}[!h]
\centering
\includegraphics[width=\columnwidth]{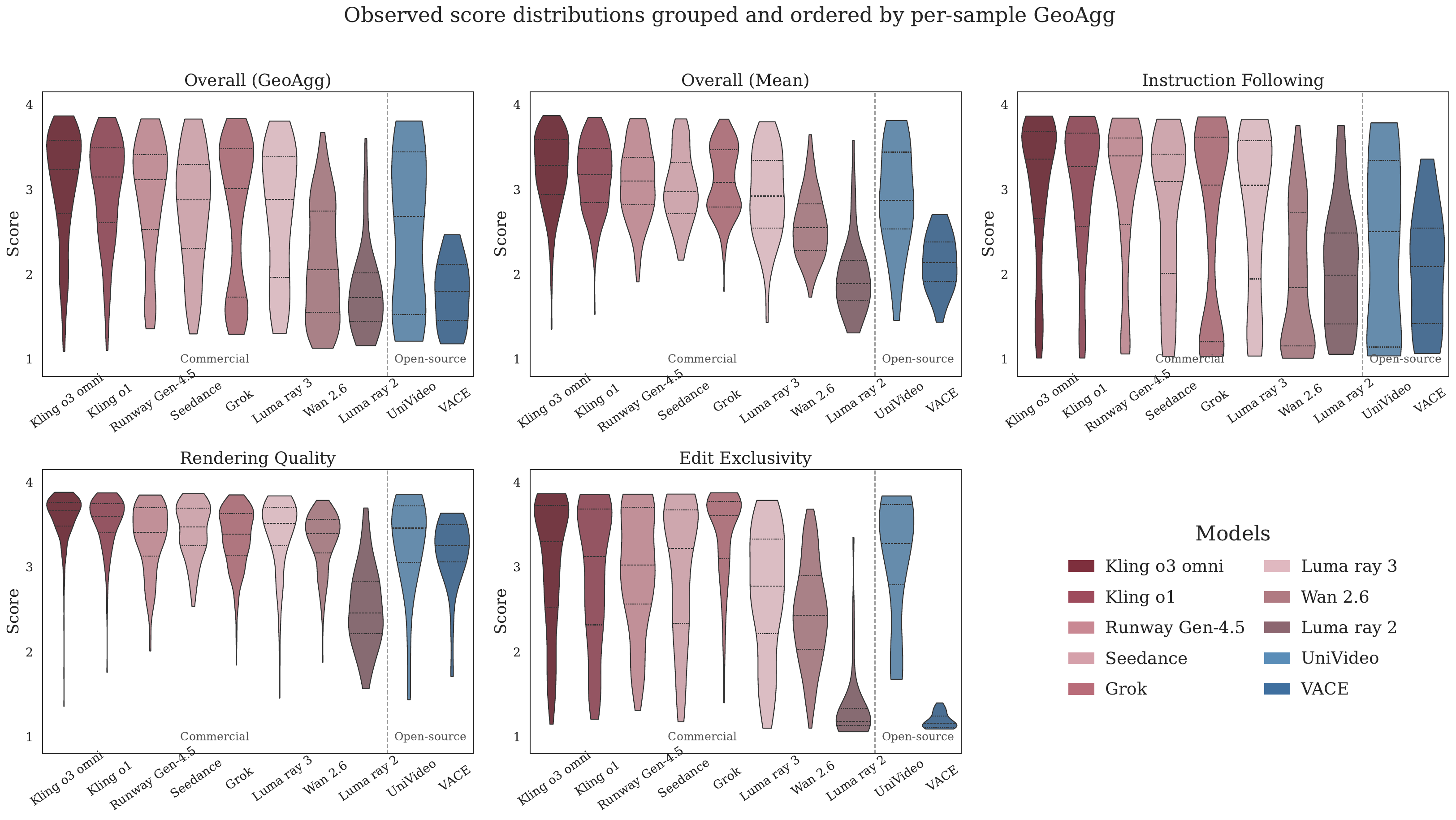}
\caption{Violin plot of the six-model evaluation, showing \GRl score density for each model and dimension.}
\label{fig:benchmark_violin_appendix}
\end{figure}

We supplement \Cref{sec:benchmark_existing_models} with two additional views of the six-model evaluation. The strip plot in \Cref{fig:benchmark_strip_appendix} shows the distribution of individual \GRl scores, while the violin plot in \Cref{fig:benchmark_violin_appendix} emphasizes the density shape for each model-dimension pair. These visualizations provide a more detailed view of score spread and concentration, complementing the main-text box plot without repeating the same summary statistics.

\section{Ethical Considerations}
\label{sec:appendix_ethics}

\paragraph{Annotator welfare.} All annotators were compensated at fair market rates and were not exposed to harmful or disturbing content. The annotation task involved evaluating video editing quality, which does not inherently involve sensitive content. NSFW content was removed during the data curation stage.

\paragraph{Potential misuse.} \GR is designed to evaluate video editing quality. While the model could theoretically be used to optimize for high-scoring edits that game specific metrics, the multi-dimensional scoring design mitigates this risk by requiring simultaneous high performance across orthogonal dimensions. The ordinal nature of the output, discrete scores 1--4, further limits the ability to exploit continuous optimization against the reward model.

\end{document}